\documentclass[a4paper,10pt]{article}
\usepackage{amsmath,amsfonts,amssymb,amsthm}
\usepackage{graphicx,subfigure,booktabs,multirow}
\usepackage{mathrsfs,enumerate,color}%,mathdots
\usepackage{algorithm,algpseudocode}
\usepackage[textwidth=14cm]{geometry}
\usepackage[sort&compress, square, numbers]{natbib}
\usepackage{hyperref}

\usepackage{diagbox}
\usepackage{threeparttable}
\usepackage{booktabs}

\newtheorem{theorem}{Theorem}[section]
\newtheorem{lemma}[theorem]{Lemma}

\newtheorem{remark}{Remark}[section]
\newtheorem{definition}{Definition}[section]
\newtheorem{example}{Example}[section]

\title{Quasi Non-Negative Quaternion  Matrix Factorization with Application to Color Face Recognition}

\author{Yifen Ke\footnote{School of Mathematics and Statistics \& FJKLMAA, Fujian Normal University,  Fuzhou  350117, People's Republic of China, and
Center for Applied Mathematics of Fujian province (FJNU), Fuzhou 350117, People's Republic of China  ({\tt keyifen@fjnu.edu.cn}).}, 
~~Changfeng Ma\footnote{School of Mathematics and Statistics \& FJKLMAA, Fujian Normal University,  Fuzhou  350117, People's Republic of China, and
Center for Applied Mathematics of Fujian province (FJNU), Fuzhou 350117, People's Republic of China ({\tt macf@fjnu.edu.cn}).},  
~~ Zhigang Jia\footnote{School of Mathematics and Statistics, Jiangsu Normal University, Xuzhou 221116, People's Republic of China, and
The Research Institute of Mathematical Science, Jiangsu Normal University, Xuzhou 221116, People's Republic of China ({\tt zhgjia@jsnu.edu.cn}).}  
~~Yajun Xie\footnote{School of Big Data, Fuzhou University of International Studies and Trade, Fuzhou 350202,  People's Republic of China  ({\tt xyj@fzfu.edu.cn}).}
~~and Riwei Liao\footnote{College of Photonic and Electronic Engineering, Fujian Normal University,  Fuzhou  350117, People's Republic of China ({\tt riweiliao2020@fjnu.edu.cn}).}
}

\begin{document}
\date{}
\maketitle
\begin{abstract}
To address the non-negativity dropout problem of  quaternion models,
a  novel quasi non-negative quaternion  matrix factorization (QNQMF) model
is presented  for color image processing.
To implement QNQMF,  the quaternion projected gradient algorithm and the quaternion alternating direction method of multipliers are proposed via formulating QNQMF as the non-convex  constraint quaternion optimization problems.
Some  properties of the  proposed  algorithms are studied.
The numerical experiments on the color image  reconstruction show that these algorithms encoded on the quaternion
perform better than these algorithms encoded on the red, green and blue channels.
Furthermore, we apply the proposed algorithms to the color face recognition.
Numerical results indicate that the accuracy rate of face recognition on the quaternion model
is better than on the red, green and blue channels of color image as well as single channel of gray level images  for the same data, when large facial expressions and shooting angle variations are presented.\\

\end{abstract}

\section{Introduction}

Color plays an important role in image processing tasks, especially in complicated scenes. The color principal component analysis \cite{QPCA_Lancos} preserves  the messages between channels  as well as the important low-frequency information. However, it is expensive to compute the  dominant eigenvectors of quaternion covariance matrix of large-scale sizes  and the non-negativity of pixel values are not well preserved.
In this paper, we present a new quasi non-negative quaternion  matrix factorization  model with non-negative constraints on the imaginary parts of quaternion matrix factors, and propose two solvers: the quaternion projected gradient algorithm and the quaternion alternating direction method of multipliers. The proposed model and numerical algorithms are successfully applied to the color face recognition, and numerical results demonstrate its efficiency and robustness,  particularly in large-scale size faces,
large facial expressions and shooting angle variations.

For given color and gray level images, the objects in color images can be recognized easier and faster than their counterparts in gray images. Recently,  it has been evinced that color does contribute to face recognizing especially when shape cues of the images are degraded \cite{Q_PCA1}.
In \cite{ADMM_color}, Rajapakse,  Tan and Rajapakse used the non-negative matrix factorization (NMF) model to the color face recognizing, which encoded separately the red, green, blue  channels and  projected  together these
feature vectors to sparse subspaces.
The numerical results presented in \cite{ADMM_color} show that the  improved accuracy
of color image face recognition over gray  image face recognition while large facial expressions and illumination variations are appeared.
For the color images, it is a common approach to treat separately for the red, green and blue channels, and then
the final processed image is obtained by combining the results from the color channels.
The downside of this approach is that the messages between channels are often ignored.
It is obvious that we can use a third-order tensor to represent the red, green and blue channels for the color image \cite{tensor_NTF}.
Each slice of this third-order tensor corresponds to one channel of the color image.
In general, it can consider the non-negative tensor factorization (NTF) to complete the low-rank and non-negative approximation.
However, the theory for NTF is not well-established when it comes to the tensor rank and inverse  for  the general tensor.

Recently, the quaternion matrix representation is playing an increasingly important role in the color image processing problems.  Pure quaternion matrices are used to represent  color images, that is,  embedding  the red, green and blue channel matrices into three imaginary parts and keeping the real part zero. Based on this, there are many successful models, such as the quaternion principal component analysis for  color face recognition \cite{Q_PCA,Q_PCA1},
 the quaternion Schur decomposition for  color watermarking \cite{Q_watermarking,Q_watermarking2},
 the quaternion singular value decomposition for color image inpainting \cite{Chen2020,ke0,Q_inpainting1},  and so on.
For more details about quaternion matrix and its applications, we refer to \cite{ke3,ke5,ke2,ke4,bu1,bu2,bu3} and the references therein.   Unfortunately, the above quaternion decomposition models ignore a fundamental property, which is the non-negativity of the imaginary parts that representing color channels.

To address this long-standing non-negativity dropout problem of quaternion models, we propose a quasi non-negative quaternion matrix factorization (QNQMF) model.  This factorization possesses the simple form, good interpretability, small storage space, and  takes full advantage of color information  between channels for color images.
In order to implement this decomposition, we formulate QNQMF as a  quaternion optimization problem.
The most difficult thing is that the obtained quaternion optimization problem is a multi-variable coupled, non-convex and non-linear complicated optimization problem.
We propose the quaternion projected gradient algorithm and the quaternion alternating direction method of multipliers  to solve QNQMF.
Then this  factorization is applied to the color face recognition.
The improved accuracy of color image recognition indicates the  recognition
performance and robustness of the implementation on the QNQMF model for color
face images.

Recall that the non-negative matrix factorization (NMF) is proposed as  a new matrix factorization model by Lee and Seung \cite{NMF_Lee} in 1999, and has been developed well in both theory and application.
Due to the simple form, good interpretability, small storage space as well as other advantages,
NMF as a powerful tool for  data analysis has been successfully used to data mining, pattern recognition  and machine learning community \cite{NMF_Lee,PT1994}.
 NMF problem  is usually formulated as a non-convex  and non-linear constraint optimization problem.
On account of  the practical application problems involved in NMF, there exists many variants of NMF using other equivalent objective functions and additional constraints
on the low-rank and non-negative factors (e.g., sparsity, orthogonality and smoothness);  see  \cite{NMF_orthogonal,NMF_sparse,NMF_constrained}.
Many numerical algorithms have been proposed to solve NMF,
for example,
the multiplicative update  method \cite{NMF_Lee},
the alternating non-negative least squares method \cite{NMF_RALS,NMF_ALS2},
the Newton-like method \cite{NMF_Newton1},
the projected gradient method \cite{NMF_Lin},
the active set method \cite{NMF_ALS1},
the alternating direction method of multipliers \cite{NMF_ADMM1,NMF_ADMM2},
the NeNMF method \cite{NMF_NeNMF} and so on.
As far as we know, NMF has not been extended to quaternion matrices with requiring the non-negative constraints on the imaginary parts of quaternion matrix until now.

This paper is organized as follows.
In Section \ref{sec2}, we review  the basic concepts of quaternion matrices and  non-negative (real) matrix factorization.
In Section \ref{sec3}, a new  QNQMF model  is established.
To solve QNQMF model, the quaternion projected gradient algorithm and
the quaternion alternating direction method of multipliers are introduced  via formulating
QNQMF  as the non-convex and non-linear constraint optimization problems in Sections \ref{sec3}  and \ref{sec4}, respectively.
We run some numerical experiments on the color image  reconstruction as well as color face recognition
to show the numerical performance of our proposals in Section \ref{sec5}.
Finally, we offer the general conclusions of this work in Section \ref{sec6}.

Throughout this paper, we denote the real field
and the quaternion algebra by $\mathbb{R}$ and $\mathbb{Q}$, respectively.
The sets  of real  and quaternion $m\times n$ matrices are denoted by $\mathbb{R}^{m\times n}$ and $\mathbb{Q}^{m\times n}$, respectively.
Let $\mathbb{R}_+^{m\times n}$ be  the non-negative set on the real field.
We use the black letters $\mathbf{A},\mathbf{B},\mathbf{C},\ldots$ to denote the  quaternion matrices and
the letters $A,B,C,\ldots$ to denote the real matrices.
The symbol $\odot$ denotes the Hadamard product, that is $(A\odot B)_{st}=a_{st}b_{st}$ for the same size real matrices $A=(a_{st})$ and $B=(b_{st})$.
The symbol $I$ is the identity matrix and $\| \cdot \|_2$ denotes the Euclidean norm of the real vector.

\section{Preliminaries}\label{sec2}
In this section, we present several basic results of quaternion matrices and the  non-negative matrix factorization on the real field.

Quaternions, introduced by the mathematician Hamilton \cite{Q_Hamilton} in 1843, are generalizations of the real and complex numbers systems.
A quaternion $\mathbf{q}$ is defined  by
$$
\mathbf{q}=q_0+q_1 {\bf i} + q_2 {\bf j} + q_3 {\bf k},
$$
where $q_0,q_1,q_2,q_3$ are real numbers,
${\bf i}$, ${\bf j}$ and ${\bf k}$  are imaginary numbers satisfying
$${\bf i}^2={\bf j}^2={\bf k}^2=-1,\quad {\bf i}{\bf j}=-{\bf j}{\bf i}={\bf k},\quad
{\bf j}{\bf k}=-{\bf k}{\bf j}={\bf i},\quad
{\bf k}{\bf i}=-{\bf i}{\bf k}={\bf j}.
$$
Let $\mathbf{p}=p_0+p_1{\bf i}+p_2{\bf j}+p_3{\bf k}\in \mathbb{Q}$
and $\mathbf{q}=q_0+q_1{\bf i}+q_2{\bf j}+q_3{\bf k}\in \mathbb{Q}$.
The sum $\mathbf{p}+\mathbf{q}$ of $\mathbf{p}$ and $\mathbf{q}$ is
$$\mathbf{p}+\mathbf{q}=(p_0+q_0)+(p_1+q_1){\bf i}+(p_2+q_2){\bf j}+(p_3+q_3){\bf k}$$
and their multiplication $\mathbf{p}\mathbf{q}$ is
\begin{align*}
\mathbf{p}\mathbf{q}=&(p_0q_0-p_1q_1-p_2q_2-p_3q_3)+(p_0q_1+p_1q_0+p_2q_3-p_3q_2){\bf i}\\
&+(p_0q_2-p_1q_3+p_2q_0+p_3q_1){\bf j}+(p_0q_3+p_1q_2-p_2q_1+p_3q_0){\bf k}.
\end{align*}
One of the most important properties of quaternions is that  the multiplication of quaternions is noncommutative as these rules,
that is $\mathbf{p}\mathbf{q}\neq \mathbf{q}\mathbf{p}$ in general for $\mathbf{p},\mathbf{q}\in \mathbb{Q}$.
For example,
it is obvious that $\mathbf{p}\mathbf{q}\neq \mathbf{q}\mathbf{p}$ while $\mathbf{p}={\bf i}$ and $\mathbf{q}={\bf j}$.
This property extends the application field of quaternions but arises the difficulty of quaternion matrix computation.

The real part of $\mathbf{q}$ is $\mathrm{Re}\,\mathbf{q}=q_0$, and its imaginary is $\mathrm{Im}\,\mathbf{q}=q_1 {\bf i} + q_2 {\bf j} + q_3 {\bf k}$. For the convenience of description,  we denote $\mathrm{Im}_i\,\mathbf{q}=q_1$, $\mathrm{Im}_j\,\mathbf{q}=q_2$ and $\mathrm{Im}_k\,\mathbf{q}=q_3$.
When $\mathrm{Re}\,\mathbf{q}=0$,  a non-zero quaternion $\mathbf{q}$ is said to be pure imaginary.
The quaternion conjugate of $\mathbf{q}$ is denoted by
$\overline{\mathbf{q}}=\mathbf{q}^*=\mathrm{Re}\,\mathbf{q}-\mathrm{Im}\,\mathbf{q}=q_0-q_1 {\bf i} - q_2 {\bf j} - q_3 {\bf k}.$
The modulus of $\mathbf{q}$ is defined by
$|\mathbf{q}|=\sqrt{\mathbf{q}\overline{\mathbf{q}}}=\sqrt{\overline{\mathbf{q}}\mathbf{q}}=\sqrt{q_0^2+q_1^2+q_2^2+q_3^2}.$
If $\mathbf{q}\neq0$, then $\mathbf{q}^{-1}=\frac{\overline{\mathbf{q}}}{|\mathbf{q}|^2}$.
We refer to \cite{Jia2019}  for more details about quaternions.

A quaternion matrix $\mathbf{A}\in\mathbb{Q}^{m\times n}$ can be denoted as
$$
\mathbf{A}=A_0+A_1 {\bf i} + A_2 {\bf j} + A_3 {\bf k},
\quad A_0, A_1, A_2 , A_3\in \mathbb{R}^{m\times n}.$$
The transpose of $\mathbf{A}$ is
$\mathbf{A}^T=A_0^T+A_1^T {\bf i} + A_2^T {\bf j} + A_3^T {\bf k}.$
The conjugate of $\mathbf{A}$ is
$\overline{\mathbf{A}}=A_0-A_1{\bf i}-A_2{\bf j}-A_3{\bf k}.$
The conjugate transpose of $\mathbf{A}$ is
$\mathbf{A}^*=A_0^T-A_1^T {\bf i}- A_2^T {\bf j}- A_3^T {\bf k}.$

We remark that the identity quaternion matrix is the same as the
classical identity matrix. Thus, we also denote as $I$.
A quaternion matrix $\mathbf{A}\in\mathbb{Q}^{n\times n}$ is inverse if there exists $\mathbf{B}\in\mathbb{Q}^{n\times n}$ such that
$\mathbf{A}\mathbf{B}=\mathbf{B}\mathbf{A}=I_n$.
And the inverse matrix of $\mathbf{A}$ is denoted by $\mathbf{A}^{-1}=\mathbf{B}$.

For $\mathbf{A}=A_0+A_1 {\bf i} + A_2 {\bf j} + A_3 {\bf k}=(\mathbf{a}_{st}), \mathbf{B}=B_0+B_1 {\bf i} + B_2 {\bf j} + B_3 {\bf k}=(\mathbf{b}_{st}) \in \mathbb{Q}^{m\times n}$,
their inner product is defined according to \cite{Q_inpainting1} by
$$\langle \mathbf{A},\mathbf{B}\rangle=\mathrm{Tr}(\mathbf{B}^*\mathbf{A})=\sum_{s=1}^m\sum_{t=1}^n\mathbf{b}_{st}^*\mathbf{a}_{st}\in \mathbb{Q},$$
where $\mathrm{Tr}(\mathbf{B}^*\mathbf{A})$ denotes the trace of $\mathbf{B}^*\mathbf{A}$.
In addition, it holds that
$$\mathrm{Re}\,\langle \mathbf{A},\mathbf{B}\rangle=\mathrm{Re}\,\langle \mathbf{B},\mathbf{A}\rangle=\langle A_0,B_0\rangle+\langle A_1,B_1\rangle+\langle A_2,B_2\rangle+\langle A_3,B_3\rangle.$$
The Frobenius norm of $\mathbf{A}=A_0+A_1 {\bf i} + A_2 {\bf j} + A_3 {\bf k}=(\mathbf{a}_{st})\in \mathbb{Q}^{m\times n}$ is defined according to \cite{Q_inpainting1} by
\begin{equation}\label{e:FnormQ}
\begin{array}{lll}
% \nonumber % Remove numbering (before each equation)
&&\|\mathbf{A}\|_F=\sqrt{\langle \mathbf{A},\mathbf{A}\rangle}=\sqrt{\mathrm{Tr}(\mathbf{A}^*\mathbf{A})}=\sqrt{\sum_{s=1}^m\sum_{t=1}^n |\mathbf{a}_{st}|^2}\\
&&=\sqrt{\langle A_0,A_0\rangle+\langle A_1,A_1\rangle+\langle A_2,A_2\rangle+\langle A_3,A_3\rangle}.
\end{array}
\end{equation}
For a quaternion matrix $\mathbf{A}=A_0+A_1 {\bf i} + A_2 {\bf j} + A_3 {\bf k}\in \mathbb{Q}^{m\times n}$, its real representation is of the following form
$$\Upsilon_\mathbf{A} =
\left(
  \begin{array}{rrrr}
    A_0 & A_2 & A_1 & A_3 \\
    -A_2 & A_0 & A_3 & -A_1 \\
    -A_1 & -A_3 & A_0 & A_2 \\
    -A_3 & A_1 & -A_2 & A_0 \\
  \end{array}
\right)\in \mathbb{R}^{4m\times 4n}.
$$
It is easy to verify that $\Upsilon_{\mathbf{A}\mathbf{B}^*}=\Upsilon_\mathbf{A} (\Upsilon_\mathbf{B})^T$ for any $\mathbf{A}\in \mathbb{Q}^{m\times l}$ and $\mathbf{B}\in \mathbb{Q}^{n\times l}$. We refer to \cite{Jia2019,Q_inpainting1}  for more details about quaternion matrix.

At the end of this section, we recall the non-negative matrix factorization \cite{NMF_Lee} on the real field. Recall that a real matrix $X$ is non-negative if and only if  all entries of $X$ are non-negative, denoted by $X\geq 0$.

{\bf Non-Negative Matrix Factorization (NMF) \cite{NMF_Lee}.}
For a given real matrix ${X}\in \mathbb{R}^{m\times n}$ with ${X}\geq 0$ and a pre-specified positive integer $l<\min(m,n)$, it finds two non-negative matrices ${W}\in \mathbb{R}^{m\times l}$ and
${H}\in \mathbb{R}^{l\times n}$ such that
\begin{equation}\label{eqa1}
{X}={W}{H}.
\end{equation}

The multiplicative update (MU) method is originally used to solve the NMF problem (\ref{eqa1}), which was first proposed by Lee and Seung in \cite{NMF_Lee} as follows
$$
\left\{
  \begin{array}{l}
({W}_{r+1})_{st} =({W}_r)_{st} \frac{({X}{H}_{r}^T)_{st}}{({W}_r{H}_{r}{H}_{r}^T)_{st}},          \\
({H}_{r+1})_{tk} =({H}_r)_{tk} \frac{({W}_{r+1}^T{X})_{tk}}{({W}_{r+1}^T{W}_{r+1}{H}_{r+1})_{tk}}   \\
  \end{array}
\right.
$$
for $s=1,2,\ldots,m$, $t=1,2,\ldots,l$ and $k=1,2,\ldots,n$.
Here, $W_r$ and $H_r$ denote the solutions at iterate $r$.
It can see that  the non-negativity of the matrices $W$ and $H$ is guaranteed
during each iteration. However,  its convergence speed is slow and the solution
obtained by MU method is not  necessarily a stationary point.

\section{Quasi non-negative quaternion  matrix factorization}\label{sec3}
In this section, a new quasi non-negative quaternion  matrix factorization model is presented and its corresponding quaternion optimization problem is discussed.

\begin{definition}
A quaternion matrix $\mathbf{Q}=Q_0+Q_1 {\bf i} + Q_2 {\bf j} + Q_3 {\bf k}\in \mathbb{Q}^{m\times n}$ is
called the quasi non-negative quaternion  matrix if $Q_1$, $Q_2$ and $Q_3$ are real non-negative matrices, that is
$$Q_1\geq 0,\quad Q_2\geq 0,\quad Q_3\geq 0.$$
The set of quasi non-negative quaternion matrices is denoted by $\mathbb{Q}_\dag^{m\times n}$.
\end{definition}

{\bf Quasi Non-Negative Quaternion  Matrix Factorization (QNQMF).}
For a given quaternion  matrix
$\mathbf{X}=X_0+X_1 {\bf i} + X_2 {\bf j} + X_3 {\bf k}\in \mathbb{Q}_\dag^{m\times n}$,
it finds the matrices $\mathbf{W}=W_0+W_1 {\bf i} + W_2 {\bf j} + W_3 {\bf k}\in \mathbb{Q}_\dag^{m\times l}$ and
$\mathbf{H}=H_0+H_1 {\bf i} + H_2 {\bf j} + H_3 {\bf k}\in \mathbb{Q}_\dag^{l\times n}$ such that
\begin{equation}\label{eqb1}
\mathbf{X}=\mathbf{W}\mathbf{H},
\end{equation}
that is
\begin{eqnarray*}
&&X_0+X_1 {\bf i} + X_2 {\bf j} + X_3 {\bf k}\\
&&=(W_0H_0-W_1H_1-W_2H_2-W_3H_3)+(W_0H_1+W_1H_0+W_2H_3-W_3H_2){\bf i}\\
&&~~+(W_0H_2-W_1H_3+W_2H_0+W_3H_1){\bf j}+(W_0H_3+W_1H_2-W_2H_1+W_3H_0){\bf k}.
\end{eqnarray*}
Here, $l$ is a pre-specified positive integer with $l< \min(m,n)$.
The matrix $\mathbf{W}$ is called the source matrix and the matrix $\mathbf{H}$ is called the activation matrix.

We take a $4$-by-$4$  quasi non-negative quaternion  matrix for illustration.
Let $\mathbf{X}=X_0+X_1\mathbf{i} +X_2\mathbf{j}+X_3\mathbf{k}$  with
$$
\begin{array}{ll}
X_0=
\left(
  \begin{array}{rrrr}
    -6  &   3  &  -2  &  -9\\
     2  &   9  &   2  &  -5\\
    -5  &   1  &  -3  &  -7\\
    -4  &   7  &   0  & -11\\
  \end{array}
\right), &
X_1=
\left(
  \begin{array}{cccc}
     3  &   3   &  7  &   3\\
     4  &   2   &  6  &   2\\
     0  &   2   &  4  &   0\\
     2  &   0   &  8  &   4\\
  \end{array}
\right),\\
X_2=
\left(
  \begin{array}{cccc}
     9 &   10  &   5  &   0\\
     8 &    4  &   2  &   4\\
     6 &    6  &   4  &   0\\
    14 &   12  &   8  &   4\\
  \end{array}
\right), &
X_3=
\left(
  \begin{array}{cccc}
     2   &  5   &  0   &  1\\
     4   &  3   &  4   &  5\\
     3   &  6   &  1   &  0\\
     2   &  7   &  2   &  1\\
  \end{array}
\right).
\end{array}
$$
Then there exist two quasi non-negative quaternion matrices $\mathbf{W}=W_0+W_1\mathbf{i} +W_2\mathbf{j}+W_3\mathbf{k}\in \mathbb{Q}^{4\times 1}$ and  $\mathbf{H}=H_0+H_1\mathbf{i} +H_2\mathbf{j}+H_3\mathbf{k}\in \mathbb{Q}^{1\times 4}$ such that
$\mathbf{X}=\mathbf{W}\mathbf{H}$, in which
\begin{eqnarray*}
% \nonumber % Remove numbering (before each equation)
W_0=
\left(
  \begin{array}{c}
     2\\
     3\\
     1\\
     3\\
  \end{array}
\right),\quad
W_1=
\left(
  \begin{array}{c}
     1\\
     0\\
     1\\
     0\\
  \end{array}
\right),\quad
W_2=
\left(
  \begin{array}{c}
     2\\
     0\\
     1\\
     2\\
  \end{array}
\right),\quad
W_3=
\left(
  \begin{array}{c}
     2\\
     1\\
     2\\
     3\\
  \end{array}
\right)
\end{eqnarray*}
and
$$
\begin{array}{ll}
H_0=\left(
      \begin{array}{cccc}
         1  &   3   &  1   &  -1 \\
      \end{array}
    \right), &
H_1=\left(
      \begin{array}{cccc}
         2  &   1   &  2   &  1 \\
      \end{array}
    \right),\\
H_2=\left(
      \begin{array}{cccc}
        2   &  1   &  0  &   1 \\
      \end{array}
    \right), & 
H_3=\left(
      \begin{array}{cccc}
        1   &  0   &  1  &   2 \\
      \end{array}
    \right).
\end{array}
$$
Here, $X_0$, $W_0$ and $H_0$ can contain negative elements. Note that QNQMF only the three imaginary parts are required to be non-negative, which greatly enriches the range of feasible factors.

\begin{remark}
In \cite{Flamant2020}, Flamant, Miron and Brie considered the following model:
$$\mathbf{X}=\mathbf{W}H,$$
where $\mathbf{X}\in  \mathbb{Q}^{m\times n}_{\mathcal{S}}$, $\mathbf{W}\in  \mathbb{Q}^{m\times l}_{\mathcal{S}}$ and $H\in \mathbb{R}_+^{l\times n}$.
Here, the set $\mathbb{Q}^{m\times n}_{\mathcal{S}}$ is defined as
$$\mathbb{Q}^{m\times n}_{\mathcal{S}}=\{\mathbf{A}=(\mathbf{a}_{st})\in \mathbb{Q}^{m\times n}~|~\mathrm{Re}(\mathbf{a}_{st})\geq 0,\quad
|\mathrm{Im}(\mathbf{a}_{st})|^2\leq \mathrm{Re}(\mathbf{a}_{st})^2\}.$$
It is obvious that the QNQMF model \eqref{eqb1} is different from the above model.

\end{remark}

\begin{remark}
The advantages of the QNQMF model \eqref{eqb1} are as follows:

$(1)$ this factorization possesses the simple form, good interpretability and small storage
space;

$(2)$ it takes full advantage of color information between channels for color images when pure quaternion matrices are used to represent color images.

\end{remark}

\subsection{Quaternion optimization problem}
To solve QNQMF (\ref{eqb1}), we consider  the following quaternion optimization problem:
\begin{equation}\label{eqb4}
\begin{array}{l}
\min~f(\mathbf{W},\mathbf{H})=\frac{1}{2}\|\mathbf{X}-\mathbf{W}\mathbf{H}\|_F^2,\\
\mathrm{s.t.} ~~~\mathbf{W}\in \mathbb{Q}_\dagger^{m\times l},~~\mathbf{H}\in \mathbb{Q}_\dagger^{l\times n}. 
\end{array}
\end{equation}
Clearly, the solution of (\ref{eqb4}) is exactly the solution of (\ref{eqb1}) if the objection function achieves zero.  So the core work of solving the QNQMF becomes to handle the quaternion optimization problem (\ref{eqb4}), which provides room for the development of new optimization methods.

The object function $f(\mathbf{W},\mathbf{H})$ is a real-valued function of two quaternion variables. According to  \cite{Chen2020},  the gradient with respect to each  quaternion variable is defined by
\begin{eqnarray}\label{eqb5W}
\nabla_\mathbf{W} f(\mathbf{W},\mathbf{H})&=&\frac{\partial f(\mathbf{W},\mathbf{H})}{\partial W_0}+\frac{\partial f(\mathbf{W},\mathbf{H})}{\partial W_1}{\bf i}
+\frac{\partial f(\mathbf{W},\mathbf{H})}{\partial W_2}{\bf j}+\frac{\partial f(\mathbf{W},\mathbf{H})}{\partial W_3}{\bf k},\\
\label{eqb5H} \nabla_\mathbf{H} f(\mathbf{W},\mathbf{H})&=&\frac{\partial f(\mathbf{W},\mathbf{H})}{\partial H_0}+\frac{\partial f(\mathbf{W},\mathbf{H})}{\partial H_1}{\bf i}
+\frac{\partial f(\mathbf{W},\mathbf{H})}{\partial H_2}{\bf j}+\frac{\partial f(\mathbf{W},\mathbf{H})}{\partial H_3}{\bf k}.
\end{eqnarray}

From the Karush-Kuhn-Tucker (KKT) optimality conditions, it is not difficult to prove that
$(\widehat{\mathbf{W}},\widehat{\mathbf{H}})$
is a stationary point of (\ref{eqb4}) if and only if
\begin{equation}\label{eqb6}
\left\{
  \begin{array}{l}
\mathrm{Re}\,\nabla_\mathbf{W} f(\widehat{\mathbf{W}},\widehat{\mathbf{H}})=0,\quad \widehat{\mathbf{W}}\in \mathbb{Q}_\dag^{m\times l},\quad \nabla_\mathbf{W} f(\widehat{\mathbf{W}},\widehat{\mathbf{H}})\in \mathbb{Q}_\dag^{m\times l},\\
\mathrm{Re}\,\nabla_\mathbf{H} f(\widehat{\mathbf{W}},\widehat{\mathbf{H}})=0, \quad \widehat{\mathbf{H}}\in \mathbb{Q}_\dag^{m\times l},\quad \nabla_\mathbf{H} f(\widehat{\mathbf{W}},\widehat{\mathbf{H}})\in \mathbb{Q}_\dag^{l\times n},\\
\mathrm{Im}_i\,\widehat{\mathbf{W}}\odot \mathrm{Im}_i\,\nabla_\mathbf{W} f(\widehat{\mathbf{W}},\widehat{\mathbf{H}})=0,\quad \mathrm{Im}_i\,\widehat{\mathbf{H}}\odot \mathrm{Im}_i\,\nabla_\mathbf{H} f(\widehat{\mathbf{W}},\widehat{\mathbf{H}})=0,\\
\mathrm{Im}_j\,\widehat{\mathbf{W}}\odot \mathrm{Im}_j\,\nabla_\mathbf{W} f(\widehat{\mathbf{W}},\widehat{\mathbf{H}})=0,\quad\mathrm{Im}_j\,\widehat{\mathbf{H}}\odot \mathrm{Im}_j\,\nabla_\mathbf{H} f(\widehat{\mathbf{W}},\widehat{\mathbf{H}})=0,\\
\mathrm{Im}_k\,\widehat{\mathbf{W}}\odot \mathrm{Im}_k\,\nabla_\mathbf{W} f(\widehat{\mathbf{W}},\widehat{\mathbf{H}})=0,\quad\mathrm{Im}_k\,\widehat{\mathbf{H}}\odot \mathrm{Im}_k\,\nabla_\mathbf{H} f(\widehat{\mathbf{W}},\widehat{\mathbf{H}})=0,\\
  \end{array}
\right.
\end{equation}
where $\mathrm{Im}_i, \mathrm{Im}_j$ and $\mathrm{Im}_k$ represent three imaginary parts of quaternion matrix.   The proof of this assertion will be given in Appendix \ref{sec:appendixB}.

The above analysis inspires us to  develop the quaternion projected gradient algorithm to handle (\ref{eqb4}).
Before this, we need provide four important lemmas, which will be used to prove the convergence of these algorithms.

From the definitions  in (\ref{e:FnormQ}), (\ref{eqb5W}) and  (\ref{eqb5H}),  the expressions of the gradient with respect to $\mathbf{W}$ and $\mathbf{H}$ are derived for the object function in  the quaternion optimization problem (\ref{eqb4}).

\begin{lemma}\label{lem3-2}
Let the real-valued function $f$ be defined as in (\ref{eqb4}).
Then we have
$$
\left\{
  \begin{array}{l}
\nabla_{\mathbf{W}} f(\mathbf{W},\mathbf{H})=-(\mathbf{X}-\mathbf{W}\mathbf{H})\mathbf{H}^*,\\
\nabla_{\mathbf{H}} f(\mathbf{W},\mathbf{H})=-\mathbf{W}^*(\mathbf{X}-\mathbf{W}\mathbf{H}).\\
  \end{array}
\right.
$$
\end{lemma}

In the KKT  optimality conditions, the equations in (\ref{eqb6}) can be further simplified.

\begin{lemma}\label{lem3-3}
The point $(\widehat{\mathbf{W}},\widehat{\mathbf{H}})\in  \mathbb{Q}_\dag^{m\times l}\times\mathbb{Q}_\dag^{l\times n} $ is the  stationary point  of (\ref{eqb4}) if and only if
\begin{equation}\label{eqb7}
\left\{
  \begin{array}{ll}
\mathrm{Re}\, [\langle \nabla_\mathbf{W} f(\widehat{\mathbf{W}},\widehat{\mathbf{H}}),~\mathbf{Y}-\widehat{\mathbf{W}}\rangle]\geq 0,  & \forall~\mathbf{Y}\in \mathbb{Q}_\dag^{m\times l}, \\
\mathrm{Re}\, [\langle \nabla_\mathbf{H} f(\widehat{\mathbf{W}},\widehat{\mathbf{H}}),~\mathbf{Z}-\widehat{\mathbf{H}}\rangle]\geq 0,  & \forall~\mathbf{Z}\in \mathbb{Q}_\dag^{l\times n}.
  \end{array}
\right.
\end{equation}
\end{lemma}

Let $\mathcal{P}_{\mathbb{Q}_\dag^{m\times n}}$ be the projection into $\mathbb{Q}_\dag^{m\times n}$. According to the Frobenius norm defined in the quaternion field,
it follows that
$$\mathcal{P}_{\mathbb{Q}_\dag^{m\times n}}(\mathbf{Q})=Q_0+\mathcal{P}_+(Q_1) {\bf i} + \mathcal{P}_+(Q_2) {\bf j} + \mathcal{P}_+(Q_3) {\bf k}\in \mathbb{Q}_\dag^{m\times n},$$
where $\mathcal{P}_+(Q_s)=\max(Q_s,0)$ for $s=1,2,3$.
Then we  obtain the properties of quasi non-negative quaternion matrix projections.

\begin{lemma}\label{lem3-5}
$(1)$ If $\mathbf{Z}\in \mathbb{Q}_\dag^{m\times n}$, then
$$\mathrm{Re}\,[\langle \mathcal{P}_{\mathbb{Q}_\dag^{m\times n}}(\mathbf{Y})-\mathbf{Y},~
\mathbf{Z}-\mathcal{P}_{\mathbb{Q}_\dag^{m\times n}}(\mathbf{Y})\rangle ]\geq 0,\quad \forall~\mathbf{Y}\in \mathbb{Q}^{m\times n}.$$

$(2)$
For any quaternion matrices $\mathbf{Y},\mathbf{Z}\in \mathbb{Q}^{m\times n}$, it has
$$\mathrm{Re}\,[\langle \mathcal{P}_{\mathbb{Q}_\dag^{m\times n}}(\mathbf{Y})-
\mathcal{P}_{\mathbb{Q}_\dag^{m\times n}}(\mathbf{Z}),~\mathbf{Y}-\mathbf{Z}\rangle ]\geq 0.$$
If $\mathcal{P}_{\mathbb{Q}_\dag^{m\times n}}(\mathbf{Y})\neq \mathcal{P}_{\mathbb{Q}_\dag^{m\times n}}(\mathbf{Z})$, then the strict inequality holds.

$(3)$ For any quaternion matrices $\mathbf{Y},\mathbf{Z}\in \mathbb{Q}^{m\times n}$, it has
$$\|\mathcal{P}_{\mathbb{Q}_\dag^{m\times n}}(\mathbf{Y})-\mathcal{P}_{\mathbb{Q}_\dag^{m\times n}}(\mathbf{Z})\|_F\leq \|\mathbf{Y}-\mathbf{Z}\|_F.$$
\end{lemma}

The proofs of above three lemmas are left in Appendix \ref{sec:appendixA}.

\subsection{Quaternion projected gradient algorithm}\label{sec4.2}
To solve the optimization problem (\ref{eqb4}), we can optimize the object function with respect to one variable  with fixing other quaternion variables:
\begin{equation}\label{eqc1}
\left\{
  \begin{array}{l}
\mathbf{W}_{r+1}\leftarrow \mathrm{argmin}_{\mathbf{W}\in \mathbb{Q}_\dag^{m\times l}}\,\frac{1}{2}\|\mathbf{X}-\mathbf{W}\mathbf{H}_{r}\|_F^2, \\
\mathbf{H}_{r+1}\leftarrow \mathrm{argmin}_{\mathbf{H}\in \mathbb{Q}_\dag^{l\times n}}\,\frac{1}{2}\|\mathbf{X}-\mathbf{W}_{r+1}\mathbf{H}\|_F^2,\\
  \end{array}
\right.
\end{equation}
where  $\mathbf{W}_{r}$ and $\mathbf{H}_{r}$ denote the solutions at iterate $r$.

%\medskip

Now, we consider a new quaternion projected gradient algorithm,
which takes advantage of the projected gradient algorithm and the simple and effective Armijo line search for the step size \cite{D1976}.

%\noindent\rule[-10pt]{13 cm}{0.1em}
\begin{algorithm}{\bf  Algorithm 1~~Quaternion projected gradient algorithm for QNQMF (\ref{eqb1})}
\label{alg3}~

\vspace{-6mm}
\noindent\rule[-10pt]{13cm}{0.05em}

{Step 0.} Given $\mathbf{X}\in \mathbb{Q}_\dag^{m\times n}$, $0 < \rho < 1$ and $0 <\sigma<1$. Initialize any feasible $\mathbf{W}_0\in \mathbb{Q}_\dag^{m\times l}$ and $\mathbf{H}_0\in \mathbb{Q}_\dag^{l\times n}$. Set $r:=0$.

{Step 1.}
Compute
$$\nabla_\mathbf{W}f(\mathbf{W}_r,\mathbf{H}_r)=-(\mathbf{X}-\mathbf{W}_r\mathbf{H}_{r})\mathbf{H}_{r}^*.$$
Update
$$\mathbf{W}_{r+1}=\mathcal{P}_{\mathbb{Q}_\dag^{m\times l}}[\mathbf{W}_r-\alpha_r \nabla_\mathbf{W}f(\mathbf{W}_r,\mathbf{H}_r)],$$
where $\alpha_r=\rho^{s_r}$ and $s_r$ is the smallest non-negative integer $s$ for which
\begin{equation}\label{eqd1}
f(\mathbf{W}_{r+1},\mathbf{H}_{r})-f(\mathbf{W}_r,\mathbf{H}_{r})\leq \sigma \mathrm{Re}\,[ \langle \nabla_\mathbf{W}f(\mathbf{W}_r,\mathbf{H}_{r}),\mathbf{W}_{r+1}-\mathbf{W}_r\rangle].
\end{equation}

{Step 2.} Compute
$$\nabla_{\mathbf{H}} f(\mathbf{W}_{r+1},\mathbf{H}_{r})=-\mathbf{W}_{r+1}^*(\mathbf{X}-\mathbf{W}_{r+1}\mathbf{H}_{r}).$$
Update
$$\mathbf{H}_{r+1}=\mathcal{P}_{\mathbb{Q}_\dag^{l\times n}}[\mathbf{H}_r-\beta_r \nabla_\mathbf{H}f(\mathbf{W}_{r+1},\mathbf{H}_r)],$$
where $\beta_r=\rho^{t_r}$ and $t_r$ is the smallest non-negative integer $t$ for which
\begin{equation}\label{eqd2}
f(\mathbf{W}_{r+1},\mathbf{H}_{r+1})-f(\mathbf{W}_{r+1},\mathbf{H}_r)\leq \sigma \mathrm{Re}\,[ \langle \nabla_\mathbf{H}f(\mathbf{W}_{r+1},\mathbf{H}_r),\mathbf{H}_{r+1}-\mathbf{H}_r\rangle].
\end{equation}

{Step 3.} If the stop termination criteria is satisfied, break; else, set $r:=r+1$, go to {Step 1}.

\end{algorithm}

%\vspace{-6mm}
%\noindent\rule[-10pt]{13cm}{0.05em}

\medskip

For Algorithm 1, we can obtain the following results.

\begin{lemma}\label{lem4.1}
In Algorithm 1, we have
\begin{eqnarray}\label{eqd3}
&&\mathrm{Re}\,[ \langle \nabla_\mathbf{W}f(\mathbf{W}_r,\mathbf{H}_{r}),\mathbf{W}_{r+1}-\mathbf{W}_r\rangle]\leq 0,\\
\label{eqd4}
&&\mathrm{Re}\,[  \langle \nabla_\mathbf{H}f(\mathbf{W}_{r+1},\mathbf{H}_r),\mathbf{H}_{r+1}-\mathbf{H}_r\rangle]\leq 0.
\end{eqnarray}
\end{lemma}

{\bf \emph {Proof}}
If $\alpha_r=0$, then $\mathbf{W}_{r+1}=\mathbf{W}_r$ and (\ref{eqd3}) holds.
If $\alpha_r>0$,
from Lemma \ref{lem3-5} (3), we have
\begin{eqnarray}
&&\mathrm{Re}\,[\langle
\mathcal{P}_{\mathbb{Q}_\dag^{m\times l}}[\mathbf{W}_r-\alpha_r \nabla_\mathbf{W}f(\mathbf{W}_r,\mathbf{H}_r)]-\mathcal{P}_{\mathbb{Q}_\dag^{m\times l}}(\mathbf{W}_r),~
\mathbf{W}_r-\alpha_r \nabla_\mathbf{W}f(\mathbf{W}_r,\mathbf{H}_r)-\mathbf{W}_r\rangle]\nonumber\\
&&=\mathrm{Re}\,[\langle
\mathbf{W}_{r+1}-\mathbf{W}_r,~-\alpha_r \nabla_\mathbf{W}f(\mathbf{W}_r,\mathbf{H}_r)\rangle]\geq 0.\nonumber
\end{eqnarray}
Hence, it follows that $\mathrm{Re}\,[\langle \nabla_\mathbf{W}f(\mathbf{W}_r,\mathbf{H}_r),~\mathbf{W}_{r+1}-\mathbf{W}_r\rangle]\leq 0 $.

Similarly, (\ref{eqd4}) is also tenable. This completes the proof.
\hfill\endproof

\medskip

\begin{lemma}\label{lem4.2}
For given $\mathbf{W},\mathbf{D_W}\in \mathbb{Q}^{m\times l}$ and $\mathbf{H},\mathbf{D_H}\in \mathbb{Q}^{l\times n}$,
the functions defined by
$$\theta(\alpha):=\frac{\|\mathcal{P}_{\mathbb{Q}_\dag^{m\times l}}(\mathbf{W}-\alpha \mathbf{D_W})-\mathbf{W}\|_F}{\alpha},\quad  \alpha>0$$
and
$$\vartheta(\beta):=\frac{\|\mathcal{P}_{\mathbb{Q}_\dag^{l\times n}}(\mathbf{H}-\beta \mathbf{D_H})-\mathbf{H}\|_F}{\beta},\quad
\beta>0$$
are nonincreasing.
\end{lemma}

{\bf \emph {Proof}} We firstly prove the following fact that if $\mathrm{Re}\,[\langle \mathbf{V},\mathbf{U}-\mathbf{V}\rangle]>0$, then
\begin{equation}\label{eqd5}
\frac{\|\mathbf{U}\|_F}{\|\mathbf{V}\|_F}\leq \frac{\mathrm{Re}\,[\langle \mathbf{U},\mathbf{U}-\mathbf{V}\rangle]}{\mathrm{Re}\,[\langle \mathbf{V},\mathbf{U}-\mathbf{V}\rangle]}
\end{equation}
for any $\mathbf{U},\mathbf{V}\in \mathbb{Q}^{m\times l}$.
In fact,
let
$$\mathbf{U}=U_0 + U_1{\bf i} + U_2{\bf j} + U_3{\bf k},\quad
\mathbf{V}=V_0 + V_1{\bf i} + V_2{\bf j} + V_3{\bf k}$$
and
$$u=\left(
    \begin{array}{c}
      \mathrm{vec}(U_0) \\
      \mathrm{vec}(U_1) \\
      \mathrm{vec}(U_2) \\
      \mathrm{vec}(U_3) \\
    \end{array}
  \right)\in \mathbb{R}^{4ml},
  \quad v=
  \left(
    \begin{array}{c}
      \mathrm{vec}(V_0) \\
      \mathrm{vec}(V_1) \\
      \mathrm{vec}(V_2) \\
      \mathrm{vec}(V_3) \\
    \end{array}
  \right)\in \mathbb{R}^{4ml},
$$
where $\mathrm{vec}(\cdot)$ is the straighten operator by column.
Note that
$$\mathrm{Re}\,[\langle \mathbf{V},\mathbf{U}\rangle]=\mathrm{Re}\,[\langle \mathbf{U},\mathbf{V}\rangle]=v^Tu,,\quad
\|\mathbf{U}\|_F^2=\|u\|_2^2,\quad
\|\mathbf{V}\|_F^2=\|v\|_2^2.$$
Based on the Cauchy-Schwarz inequality $u^Tv\leq\|u\|_2\|v\|_2$,
it follows that
$$\mathrm{Re}\,[\langle \mathbf{V},\mathbf{U}\rangle]\leq \|\mathbf{U}\|_F\|\mathbf{V}\|_F,\quad \forall\,\mathbf{U},\mathbf{V}\in \mathbb{Q}^{m\times l}.$$
If $\mathrm{Re}\,[\langle \mathbf{V},\mathbf{U}-\mathbf{V}\rangle]>0$, we have
\begin{eqnarray}
&&\frac{\|\mathbf{U}\|_F}{\|\mathbf{V}\|_F}\leq \frac{\mathrm{Re}\,[\langle \mathbf{U},\mathbf{U}-\mathbf{V}\rangle]}{\mathrm{Re}\,[\langle \mathbf{V},\mathbf{U}-\mathbf{V}\rangle]}\nonumber\\
&&\Leftrightarrow
\|\mathbf{U}\|_F \mathrm{Re}\,[\langle \mathbf{V},\mathbf{U}-\mathbf{V}\rangle]\leq
\|\mathbf{V}\|_F \mathrm{Re}\,[\langle \mathbf{U},\mathbf{U}-\mathbf{V}\rangle]\nonumber\\
&&\Leftrightarrow
\|\mathbf{U}\|_F  \mathrm{Re}\,[\langle \mathbf{V},\mathbf{U}\rangle]+
\|\mathbf{V}\|_F \mathrm{Re}\,[\langle \mathbf{U},\mathbf{V}\rangle]\leq
\|\mathbf{U}\|_F\|\mathbf{V}\|_F(\|\mathbf{U}\|_F+\|\mathbf{V}\|_F)\nonumber\\
&&\Leftrightarrow
\mathrm{Re}\,[\langle \mathbf{V},\mathbf{U}\rangle]\leq
\|\mathbf{U}\|_F\|\mathbf{V}\|_F.\nonumber
\end{eqnarray}

Next, we prove that $\theta(\alpha)$ is a nonincreasing function for $\alpha>0$.
Let $\alpha_1>\alpha_2>0$ be given.

Case 1: if $\mathcal{P}_{\mathbb{Q}_\dag^{m\times l}}(\mathbf{W}-\alpha_1 \mathbf{D_W})=\mathcal{P}_{\mathbb{Q}_\dag^{m\times l}}(\mathbf{W}-\alpha_2 \mathbf{D_W})$, then 
it has $\theta(\alpha_1)\leq \theta(\alpha_2)$.

Case 2: if $\mathcal{P}_{\mathbb{Q}_\dag^{m\times l}}(\mathbf{W}-\alpha_1 \mathbf{D_W})\neq \mathcal{P}_{\mathbb{Q}_\dag^{m\times l}}(\mathbf{W}-\alpha_2 \mathbf{D_W})$, then let
$$\mathbf{U}=\mathcal{P}_{\mathbb{Q}_\dag^{m\times l}}(\mathbf{W}-\alpha_1 \mathbf{D_W})-\mathbf{W},\quad
\mathbf{V}=\mathcal{P}_{\mathbb{Q}_\dag^{m\times l}}(\mathbf{W}-\alpha_2 \mathbf{D_W})-\mathbf{W}.
$$
From Lemma \ref{lem3-5} (1), we have
$$\mathrm{Re}\,[\langle
\mathcal{P}_{\mathbb{Q}_\dag^{m\times n}}(\mathbf{W}-\alpha_1 \mathbf{D_W})-(\mathbf{W}-\alpha_1 \mathbf{D_W}),\,
\mathcal{P}_{\mathbb{Q}_\dag^{m\times n}}(\mathbf{W}-\alpha_2 \mathbf{D_W})-\mathcal{P}_{\mathbb{Q}_\dag^{m\times n}}(\mathbf{W}-\alpha_1 \mathbf{D_W})\rangle ]\geq 0,$$
that is
\begin{eqnarray}\label{eqd6}
&&\mathrm{Re}\,[\langle
\mathbf{U},~\mathbf{V}-\mathbf{U}\rangle ]+
\alpha_1
\mathrm{Re}\,[\langle \mathbf{D_W},~\mathbf{V}-\mathbf{U}\rangle ]\geq0 \nonumber\\
&&\Leftrightarrow
\mathrm{Re}\,[\langle
\mathbf{U},~\mathbf{U}-\mathbf{V}\rangle ]\leq
\alpha_1
\mathrm{Re}\,[\langle \mathbf{D_W},~\mathbf{V}-\mathbf{U}\rangle ].
\end{eqnarray}
Again, we have
$$\mathrm{Re}[\langle
\mathcal{P}_{\mathbb{Q}_\dag^{m\times n}}(\mathbf{W}-\alpha_2 \mathbf{D_W})-(\mathbf{W}-\alpha_2 \mathbf{D_W}),\,
\mathcal{P}_{\mathbb{Q}_\dag^{m\times n}}(\mathbf{W}-\alpha_1 \mathbf{D_W})-\mathcal{P}_{\mathbb{Q}_\dag^{m\times n}}(\mathbf{W}-\alpha_2 \mathbf{D_W})\rangle]\geq 0,$$
that is
\begin{eqnarray}\label{eqd7}
&&\mathrm{Re}\,[\langle
\mathbf{V},~\mathbf{U}-\mathbf{V}\rangle ]+
\alpha_2
\mathrm{Re}\,[\langle \mathbf{D_W},~\mathbf{U}-\mathbf{V}\rangle ]\geq0\nonumber\\
&&\Leftrightarrow
\mathrm{Re}\,[\langle
\mathbf{V},~\mathbf{U}-\mathbf{V}\rangle ]\geq
\alpha_2
\mathrm{Re}\,[\langle \mathbf{D_W},~\mathbf{V}-\mathbf{U}\rangle ].
\end{eqnarray}
Since $\alpha_1>\alpha_2>0$ and $\mathcal{P}_{\mathbb{Q}_\dag^{m\times l}}(\mathbf{W}-\alpha_1 \mathbf{D_W})\neq \mathcal{P}_{\mathbb{Q}_\dag^{m\times l}}(\mathbf{W}-\alpha_2 \mathbf{D_W})$, thus
$$\mathrm{Re}\,[\langle
\mathcal{P}_{\mathbb{Q}_\dag^{m\times n}}(\mathbf{W}-\alpha_2 \mathbf{D_W})-\mathcal{P}_{\mathbb{Q}_\dag^{m\times n}}(\mathbf{W}-\alpha_1 \mathbf{D_W}),~
(\alpha_1-\alpha_2)\mathbf{D_W}\rangle ]> 0,$$
it means that
\begin{equation}\label{eqd8}
\mathrm{Re}\,[\langle \mathbf{D_W},~\mathbf{V}-\mathbf{U}\rangle ]>0.
\end{equation}
From (\ref{eqd5}), (\ref{eqd6}), (\ref{eqd7}) and (\ref{eqd8}),  it has
\begin{eqnarray}
\frac{\|\mathcal{P}_{\mathbb{Q}_\dag^{m\times l}}(\mathbf{W}-\alpha_1 \mathbf{D_W})-\mathbf{W}\|_F}{\|\mathcal{P}_{\mathbb{Q}_\dag^{m\times l}}(\mathbf{W}-\alpha_2 \mathbf{D_W})-\mathbf{W}\|_F}\leq \frac{\mathrm{Re}\,[\langle \mathbf{U},\mathbf{U}-\mathbf{V}\rangle]}{\mathrm{Re}\,[\langle \mathbf{V},\mathbf{U}-\mathbf{V}\rangle]}
\leq \frac{\alpha_1}{\alpha_2},\nonumber
\end{eqnarray}
then it has $\theta(\alpha_1)\leq \theta(\alpha_2)$.

Similarly, $\vartheta(\beta)$ is a nonincreasing function for $\beta>0$. This completes the proof.
\hfill\endproof

\medskip

For simplicity of analysis, denote
\begin{eqnarray}
&&\mathbf{W}_r(\alpha):=\mathcal{P}_{\mathbb{Q}_\dag^{m\times l}}[\mathbf{W}_r-\alpha  \nabla_\mathbf{W}f(\mathbf{W}_r,\mathbf{H}_r)], \nonumber\\
&&\mathbf{H}_r(\beta):=\mathcal{P}_{\mathbb{Q}_\dag^{l\times n}}[\mathbf{H}_r-\beta  \nabla_\mathbf{H}f(\mathbf{W}_{r+1},\mathbf{H}_r)].\nonumber
\end{eqnarray}

\begin{lemma}
Algorithm 1 is well defined.
\end{lemma}

{\bf \emph {Proof}}
We only prove that there exist $\alpha_W$ and $\beta_H$
such that
\begin{equation}\label{eqd9}
f(\mathbf{W}(\alpha),\mathbf{H})-f(\mathbf{W},\mathbf{H})\leq \sigma \mathrm{Re}\,[ \langle\nabla_\mathbf{W} f(\mathbf{W},\mathbf{H}),\mathbf{W}(\alpha)-\mathbf{W}\rangle],~~ \forall \alpha\in [0,\alpha_W]
\end{equation}
and
\begin{equation}\label{eqd10}
f(\mathbf{W},\mathbf{H}(\beta))-f(\mathbf{W},\mathbf{H})\leq \sigma \mathrm{Re}\,[ \langle\nabla_\mathbf{H} f(\mathbf{W},\mathbf{H}),\mathbf{H}(\beta)-\mathbf{H}\rangle],~~ \forall \beta\in [0,\beta_H]
\end{equation}
for given $\sigma>0$.
In other word, the stepsizes satisfying (\ref{eqd1}) and (\ref{eqd2})
will be found after  finite number. Hence,   Algorithm 1 is well defined.

In fact, if $\nabla_\mathbf{W} f(\mathbf{W},\mathbf{H})=0$, then the conclusion (\ref{eqd9}) holds with $\alpha_W$ being
any positive scalar. Now, we assume that $\nabla_\mathbf{W} f(\mathbf{W},\mathbf{H})\neq0$, that is
$\|\mathbf{W}(\alpha)-\mathbf{W}\|_F\neq 0$ for all $\alpha\in (0,1]$.
From Lemma \ref{lem3-5} (1), it has
$$\mathrm{Re}\,[\langle
\mathbf{W}(\alpha)-\mathbf{W},~\mathbf{W}-\alpha \nabla_\mathbf{W}f(\mathbf{W},\mathbf{H})-\mathbf{W}(\alpha)\rangle]\geq 0.$$
Hence, for all $\alpha\in (0,1]$, it follows Lemma \ref{lem4.2} that
\begin{equation}\label{eqd11}
  \begin{array}{ll}
&\mathrm{Re}\,[\langle \nabla_\mathbf{W}f(\mathbf{W},\mathbf{H}),~\mathbf{W}-\mathbf{W}(\alpha) \rangle] \\
&\geq\frac{\|\mathbf{W}(\alpha)-\mathbf{W}\|_F^2}{\alpha}\geq \|\mathbf{W}(1)-\mathbf{W}\|_F\,\|\mathbf{W}(\alpha)-\mathbf{W}\|_F. \\
  \end{array}
\end{equation}
By the mean value theorem, we have
\begin{eqnarray}\label{eqd12}
&&f(\mathbf{W}(\alpha),\mathbf{H})-f(\mathbf{W},\mathbf{H})\nonumber\\
&&=\mathrm{Re}\, [\langle \nabla_\mathbf{W}f(\mathbf{W}(\xi_\alpha),\mathbf{H}),~\mathbf{W}(\alpha)-\mathbf{W}\rangle]\nonumber\\
&&= \mathrm{Re}\, [\langle \nabla_\mathbf{W}f(\mathbf{W},\mathbf{H}),~\mathbf{W}(\alpha)-\mathbf{W}\rangle]\nonumber\\
&&~~+\mathrm{Re}\, [\langle
\nabla_\mathbf{W}f(\mathbf{W}(\xi_\alpha),\mathbf{H})-\nabla_\mathbf{W}f(\mathbf{W},\mathbf{H}),~\mathbf{W}(\alpha)-\mathbf{W}\rangle],
\end{eqnarray}
where $\mathbf{W}(\xi_{\alpha})$  lies in the line segment joining $\mathbf{W}$ and $\mathbf{W}(\alpha)$.
From (\ref{eqd9})  and (\ref{eqd12}), it can be written as
\begin{equation}\label{eqd13}
\begin{array}{ll}
&(1-\sigma) \mathrm{Re}\,[ \langle\nabla_\mathbf{W} f(\mathbf{W},\mathbf{H}),~\mathbf{W}-\mathbf{W}(\alpha)\rangle]\\
&\geq \mathrm{Re}\, [\langle \nabla_\mathbf{W}f(\mathbf{W},\mathbf{H})-\nabla_\mathbf{W}f(\mathbf{W}(\xi_\alpha),\mathbf{H}),~
\mathbf{W}-\mathbf{W}(\alpha)\rangle].
\end{array}
\end{equation}
Therefore, (\ref{eqd13}) is satisfied for all $\alpha\in (0,1]$ such that
$$(1-\sigma)\|\mathbf{W}(1)-\mathbf{W}\|_F \geq \mathrm{Re}\, [\langle
\nabla_\mathbf{W}f(\mathbf{W},\mathbf{H})-\nabla_\mathbf{W}f(\mathbf{W}(\xi_\alpha),\mathbf{H}),~
\frac{\mathbf{W}-\mathbf{W}(\alpha)}{\|\mathbf{W}(\alpha)-\mathbf{W}\|_F}\rangle].$$
By the continuity of norm and inner product, there exists $\alpha_W$ such that the   above relation
holds.
Therefor, (\ref{eqd13}) and (\ref{eqd9}) also hold.

For (\ref{eqd10}), it can be proved similarly.
\hfill\endproof

\medskip

\begin{lemma}\label{lem4.4}
The sequence $\{f(\mathbf{W}_r,\mathbf{H}_r)\}_{r=0}^{\infty}$ generated by Algorithm 1 is monotonically nonincreasing.
\end{lemma}

{\bf \emph {Proof}}
From Lemma \ref{lem4.1}, we have
$$f(\mathbf{W}_{r+1},\mathbf{H}_{r})-f(\mathbf{W}_r,\mathbf{H}_r)\leq 0,\quad f(\mathbf{W}_{r+1},\mathbf{H}_{r+1})-f(\mathbf{W}_{r+1},\mathbf{H}_{r})\leq 0.$$
Hence, it can get that
$$f(\mathbf{W}_{r+1},\mathbf{H}_{r+1})-f(\mathbf{W}_r,\mathbf{H}_r)\leq 0.$$
This completes the proof.
\hfill\endproof

\medskip

\begin{theorem}
Let $\{(\mathbf{W}_r,\mathbf{H}_r)\}_{r=0}^{+\infty}$ be the sequence generated by Algorithm 1.
If
$$\lim_{r\rightarrow +\infty}\mathbf{W}_r=\widehat{\mathbf{W}},\quad \lim_{r\rightarrow +\infty}\mathbf{H}_r=\widehat{\mathbf{H}},$$
then $(\widehat{\mathbf{W}},\widehat{\mathbf{H}})$ is a stationary point of QNQMF (\ref{eqb1}).
\end{theorem}

{\bf \emph {Proof}} Since $\{f(\mathbf{W}_r,\mathbf{H}_r)\}_{r=0}^{+\infty}$ is monotonically nonincreasing, then it holds that
$f(\mathbf{W}_r,\mathbf{H}_r)\rightarrow f(\widehat{\mathbf{W}},\widehat{\mathbf{H}})$ as $r\rightarrow +\infty$.
We  prove this theorem from two cases.

Case 1: $\liminf_{r\rightarrow +\infty} \alpha_r \geq \widehat{\alpha}>0$.
From (\ref{eqd1}) and (\ref{eqd11}), for the sufficiently large $r$, we have
\begin{eqnarray}
&&f(\mathbf{W}_r,\mathbf{H}_{r})-f(\mathbf{W}_{r+1},\mathbf{H}_{r})\geq
\sigma \mathrm{Re}\,[ \langle \nabla_\mathbf{W}f(\mathbf{W}_r,\mathbf{H}_{r}),\mathbf{W}_r-\mathbf{W}_{r+1}\rangle]\nonumber\\
&&\geq \frac{\sigma\alpha_r\|\mathbf{W}_{r+1}-\mathbf{W}_r\|_F^2}{\alpha_r^2}
\geq \sigma\alpha_r\|\mathcal{P}_{\mathbb{Q}_\dag^{m\times l}}(\mathbf{W}_r- \nabla_\mathbf{W}f(\mathbf{W}_r,\mathbf{H}_r))-\mathbf{W}_r\|_F^2.\nonumber
\end{eqnarray}
Taking the limit as $r\rightarrow +\infty$, we can obtain that
$$0\geq \sigma \widehat{\alpha} \|\mathcal{P}_{\mathbb{Q}_\dag^{m\times l}}(\widehat{\mathbf{W}}- \nabla_\mathbf{W}f(\widehat{\mathbf{W}},\widehat{\mathbf{H}}))-\widehat{\mathbf{W}}\|_F^2.$$
Hence, it has $\widehat{\mathbf{W}}=\mathcal{P}_{\mathbb{Q}_\dag^{m\times l}}(\widehat{\mathbf{W}}- \nabla_\mathbf{W}f(\widehat{\mathbf{W}},\widehat{\mathbf{H}}))$.

Case 2: $\liminf_{r\rightarrow +\infty} \alpha_r=0$. Then there exists a subsequence $\{\alpha_r\} _{r\in \widehat{R}}$ with the set $\widehat{R}\subseteq\{0,1,2,\ldots\}$ converging to zero.
For all $r\in \widehat{R}$, which are sufficiently large, (\ref{eqd1}) will be failed at least once
and therefore
\begin{equation}\label{eqd14}
  \begin{array}{ll}
&f(\mathbf{W}_{r}(\rho^{-1}\alpha_r),\mathbf{H}_{r})-f(\mathbf{W}_r,\mathbf{H}_{r})\\
&>\sigma \mathrm{Re}\,[ \langle \nabla_\mathbf{W}f(\mathbf{W}_r,\mathbf{H}_{r}),~\mathbf{W}_{r}(\rho^{-1}\alpha_r)-\mathbf{W}_r\rangle].
  \end{array}
\end{equation}
Hence, $\mathbf{W}_r\neq \mathbf{W}_{r}(\rho^{-1}\alpha_r)$; otherwise, it will contradict with (\ref{eqd14}).
Thus, it holds $\|\mathbf{W}_r- \mathbf{W}_{r}(\rho^{-1}\alpha_r)\|_F>0$.

By the mean value theorem, we have
\begin{eqnarray}\label{eqd15}
&&f(\mathbf{W}_r(\rho^{-1}\alpha_r),\mathbf{H}_r)-f(\mathbf{W}_r,\mathbf{H}_r)\nonumber\\
&&=\mathrm{Re}\, [\langle \nabla_\mathbf{W}f(\mathbf{W}_r(\xi_{\alpha_r}),\mathbf{H}_r),~\mathbf{W}_r(\rho^{-1}\alpha)-\mathbf{W}_r\rangle]\nonumber\\
&&= \mathrm{Re}\, [\langle \nabla_\mathbf{W}f(\mathbf{W}_r,\mathbf{H}_r),~\mathbf{W}_r(\rho^{-1}\alpha_r)-\mathbf{W}_r\rangle]\nonumber\\
&&~~+\mathrm{Re}\, [\langle
\nabla_\mathbf{W}f(\mathbf{W}_r(\xi_{\alpha_r}),\mathbf{H}_r)-\nabla_\mathbf{W}f(\mathbf{W}_r,\mathbf{H}_r),~\mathbf{W}_r(\rho^{-1}\alpha_r)-\mathbf{W}_r\rangle],
\end{eqnarray}
where $\mathbf{W}_r(\xi_{\alpha_r})$  lies in the line segment joining $\mathbf{W}_r$ and $\mathbf{W}_r(\rho^{-1}\alpha_r)$.
Combine (\ref{eqd14}) and (\ref{eqd15}), it has
\begin{eqnarray}
&&(1-\sigma)\mathrm{Re}\, [\langle \nabla_\mathbf{W}f(\mathbf{W}_r,\mathbf{H}_r),~\mathbf{W}_r-\mathbf{W}_r(\rho^{-1}\alpha_r)\rangle]\nonumber\\
&&<\mathrm{Re}\, [\langle
\nabla_\mathbf{W}f(\mathbf{W}_r(\xi_{\alpha_r}),\mathbf{H}_r)-\nabla_\mathbf{W}f(\mathbf{W}_r,\mathbf{H}_r),~\mathbf{W}_r(\rho^{-1}\alpha_r)-\mathbf{W}_r\rangle].\nonumber
\end{eqnarray}
According to Lemma \ref{lem3-5} (1) and Lemma \ref{lem4.2}, it follows that
\begin{eqnarray}
&&\mathrm{Re}\,[\langle \nabla_\mathbf{W}f(\mathbf{W}_r,\mathbf{H}_r),~\mathbf{W}_r-\mathbf{W}_r(\rho^{-1}\alpha_r) \rangle]\nonumber\\
&&\geq\frac{\|\mathbf{W}_r(\rho^{-1}\alpha_r)-\mathbf{W}_r\|_F^2}{\rho^{-1}\alpha_r}\geq \|\mathbf{W}_r(1)-\mathbf{W}_r\|_F\,\|\mathbf{W}_r(\rho^{-1}\alpha_r)-\mathbf{W}_r\|_F.\nonumber
\end{eqnarray}
Hence, it has
\begin{eqnarray*}
&&(1-\sigma)\|\mathbf{W}_r(1)-\mathbf{W}_r\|_F\,\|\mathbf{W}_r(\rho^{-1}\alpha_r)-\mathbf{W}_r\|_F\nonumber\\
&&< \mathrm{Re}\, [\langle
\nabla_\mathbf{W}f(\mathbf{W}_r(\xi_{\alpha_r}),\mathbf{H}_r)-\nabla_\mathbf{W}f(\mathbf{W}_r,\mathbf{H}_r),~\mathbf{W}_r(\rho^{-1}\alpha_r)-\mathbf{W}_r\rangle]\nonumber\\
&&\leq \|\nabla_\mathbf{W}f(\mathbf{W}_r(\xi_{\alpha_r}),\mathbf{H}_r)-\nabla_\mathbf{W}f(\mathbf{W}_r,\mathbf{H}_r)\|_F\|\mathbf{W}_r(\rho^{-1}\alpha_r)-\mathbf{W}_r\|_F.
\end{eqnarray*}
We obtain that
\begin{equation}\label{eqd16}
(1-\sigma)\|\mathbf{W}_r(1)-\mathbf{W}_r\|_F<\|\nabla_\mathbf{W}f(\mathbf{W}_r(\xi_{\alpha_r}),\mathbf{H}_r)-\nabla_\mathbf{W}f(\mathbf{W}_r,\mathbf{H}_r)\|_F.
\end{equation}
Since $\alpha_r\rightarrow 0$ and $\mathbf{W}_r\rightarrow \widehat{\mathbf{W}}$ as $r\rightarrow +\infty$ and $r\in \widehat{R}$, it follows that $\mathbf{W}_r(\xi_{\alpha_r})\rightarrow \widehat{\mathbf{W}}$.
Taking the limit in (\ref{eqd16}) as $r\rightarrow +\infty$, we can obtain
$(1-\sigma)\|\widehat{\mathbf{W}}(1)-\widehat{\mathbf{W}}\|_F\leq 0$  that is
$\widehat{\mathbf{W}}=\mathcal{P}_{\mathbb{Q}_\dag^{m\times l}}(\widehat{\mathbf{W}}- \nabla_\mathbf{W}f(\widehat{\mathbf{W}},\widehat{\mathbf{H}}))$.

From Lemma \ref{lem3-5} (1) and  $\widehat{\mathbf{W}}=\mathcal{P}_{\mathbb{Q}_\dag^{m\times l}}(\widehat{\mathbf{W}}- \nabla_\mathbf{W}f(\widehat{\mathbf{W}},\widehat{\mathbf{H}}))$, for any $\mathbf{Y}\in \mathbb{Q}_\dag^{m\times l}$, it follows that
\begin{eqnarray*}
&&\mathrm{Re}\,[\langle \mathcal{P}_{\mathbb{Q}_\dag^{m\times l}}(\widehat{\mathbf{W}}- \nabla_\mathbf{W}f(\widehat{\mathbf{W}},\widehat{\mathbf{H}}))-(\widehat{\mathbf{W}}-\nabla_\mathbf{W}f(\widehat{\mathbf{W}},\widehat{\mathbf{H}})),~
\mathbf{Y}-\mathcal{P}_{\mathbb{Q}_\dag^{m\times l}}(\widehat{\mathbf{W}}- \nabla_\mathbf{W}f(\widehat{\mathbf{W}},\widehat{\mathbf{H}})\rangle]\\
&&=\mathrm{Re}\,[\langle\widehat{\mathbf{W}}-(\widehat{\mathbf{W}}- \nabla_\mathbf{W}f(\widehat{\mathbf{W}},\widehat{\mathbf{H}})),~
\mathbf{Y}-\widehat{\mathbf{W}}\rangle]\\
&&=\mathrm{Re}\,[\langle \nabla_\mathbf{W}f(\widehat{\mathbf{W}},\widehat{\mathbf{H}}),~
\mathbf{Y}-\widehat{\mathbf{W}}\rangle]\geq 0.
\end{eqnarray*}
Similarly,  it also has
$$\mathrm{Re}\,[\langle \nabla_\mathbf{H}f(\widehat{\mathbf{W}},\widehat{\mathbf{H}}),~
\mathbf{Z}-\widehat{\mathbf{H}}\rangle ]\geq 0,\quad \forall~\mathbf{Z}\in \mathbb{Q}_\dag^{l\times n}.$$
According to Lemma \ref{lem3-3}, it gets that $(\widehat{\mathbf{W}},\widehat{\mathbf{H}})$ is the  stationary point.
This completes the proof.
\hfill\endproof

\medskip

Algorithm 1 is well defined and produces a monotonic non-increasing sequence.
In fact, Algorithm 1 may cost the most time to search the step sizes $\alpha_r$ and
$\beta_r$.  So one should check as few step sizes as possible. In practice, $\alpha_r$ and $\beta_r$ may be similar to the previous step  sizes $\alpha_{r-1}$ and $\beta_{r-1}$, it can use $\alpha_{r-1}$ and $\beta_{r-1}$ as the initial guesses and then either increases or decreases them in order to find the  largest $\alpha_r$ and $\beta_r$ satisfying (\ref{eqd1}) and (\ref{eqd2}), respectively.
Sometimes, a larger step more effectively projects variables to bounds at one
iteration. Algorithm 2 implements a better initial guesses of $\alpha_r$ and $\beta_r$ at each iteration, called
the quaternion improved projected gradient (QIPG) algorithm.

%\noindent\rule[-10pt]{13 cm}{0.1em}
\begin{algorithm}{\bf Algorithm 2~~QIPG algorithm for QNQMF (\ref{eqb1})}\label{alg4}~

\vspace{-6mm}
\noindent\rule[-10pt]{13cm}{0.05em}

{Step 0.} Given $\mathbf{X}\in \mathbb{Q}_\dag^{m\times n}$, $0 < \rho < 1$ and $0 <\sigma<1$. Initialize any feasible $\mathbf{W}_0\in \mathbb{Q}_\dag^{m\times l}$ and $\mathbf{H}_0\in \mathbb{Q}_\dag^{l\times n}$. Set $r:=0$, $\alpha_{-1}=0$ and $\beta_{-1}=0$.

{Step 1.}
$(1)$ ~Compute
$$\nabla_\mathbf{W}f(\mathbf{W}_r,\mathbf{H}_{r})=-(\mathbf{X}-\mathbf{W}_r\mathbf{H}_{r})\mathbf{H}_{r}^*.$$

$(2)$ ~Set $\alpha_r\leftarrow \alpha_{r-1}$.  If $\alpha_r$ satisfies (\ref{eqd1}), repeatedly increase it by
$$\alpha_r \leftarrow \alpha_r /\rho$$
until $\alpha_r $ does not satisfy (\ref{eqd1}).
Else, repeatedly decrease $\alpha_r$ by
$$\alpha_r \leftarrow \alpha_r\cdot \rho$$
until $\alpha_r $ satisfies (\ref{eqd1}).

$(3)$ ~Update
$$\mathbf{W}_{r+1}=\mathcal{P}_{\mathbb{Q}_\dag^{m\times l}}[\mathbf{W}_r-\alpha_r \nabla_\mathbf{W}f(\mathbf{W}_r,\mathbf{H}_{r})].$$

{Step 2.}
$(1)$ ~Compute
$$\nabla_{\mathbf{H}} f(\mathbf{W}_{r+1},\mathbf{H}_{r})=-\mathbf{W}_{r+1}^*(\mathbf{X}-\mathbf{W}_{r+1}\mathbf{H}_{r}).$$

$(2)$ ~Set $\beta_r\leftarrow \beta_{r-1}$.  If $\beta_r$ satisfies (\ref{eqd2}), repeatedly increase it by
$$\beta_r \leftarrow \beta_r /\rho$$
until  $\beta_r $ does not satisfy (\ref{eqd2}).
Else, repeatedly decrease $\beta_r$ by
$$\beta_r \leftarrow \beta_r\cdot \rho$$
until $\beta_r $ satisfies (\ref{eqd2}).

$(3)$ ~Update
$$\mathbf{H}_{r+1}=\mathcal{P}_{\mathbb{Q}_\dag^{l\times n}}[\mathbf{H}_r-\beta_r \nabla_\mathbf{H}f(\mathbf{W}_{r+1},\mathbf{H}_r)].$$

{Step 3.} If the stop termination criteria is satisfied, break; else, set $r:=r+1$, go to {Step 1}.

\end{algorithm}

%\vspace{-6mm}
%\noindent\rule[-10pt]{13cm}{0.05em}

\medskip

\begin{remark}
Algorithm 1 is a general version of the projected gradient methods for the non-negative matrix factorization (\ref{eqa1}) proposed by Lin in \cite{NMF_Lin}.
The similar theoretical results can be obtained via  that in \cite{Gafni1982,Calamai1987}.
\end{remark}

\begin{remark}
From Algorithms 1 and 2, it can see
that the non-negativity of the imaginary parts of $\mathbf{W}$ and $\mathbf{H}$ that representing color channels are  maintained during the iteration.
\end{remark}

\section{Quaternion alternating direction method of multipliers}\label{sec4}
In this section, we present a novel equivalent model of QNQMF and then solve it based on the idea of the alternating direction method of multipliers (ADMM).

The ADMM framework was originally proposed by Glowinski and Marrocco in \cite{ADMMGlowinski1975}. Recently, it becomes increasingly popular in machine learning and signal processing.
In particular, Boyd et al. \cite{ADMMBoyd2011} firstly proposed the ADMM to solve the non-convex NMF problem.

In order to utilize the ADMM framework,  we consider the following equivalent  programming
problem of QNQMF (\ref{eqb4})
\begin{equation}\label{eqe1}
\begin{array}{l}
\min~f(\mathbf{W},\mathbf{H})=\frac{1}{2}\|\mathbf{X}-\mathbf{W}\mathbf{H}\|_F^2,\\
\mathrm{s.t.} ~~~\mathbf{W}=\mathbf{U},~~\mathbf{H}=\mathbf{V},~~\mathbf{U}\in \mathbb{Q}_\dagger^{m\times l},~~\mathbf{V}\in \mathbb{Q}_\dagger^{l\times n}.
\end{array}
\end{equation}
Then, the augmented Lagrangian function of (\ref{eqe1}) is
\begin{equation}\label{eqe2}
\begin{array}{ll}
&\mathcal{L}_A(\mathbf{W},\mathbf{H},\mathbf{U},\mathbf{V},\mathbf{\Lambda},\mathbf{\Pi})=\frac{1}{2}\|\mathbf{X}-\mathbf{W}\mathbf{H}\|_F^2\\
&-\mathrm{Re}\,[\langle \mathbf{\Lambda},~\mathbf{W}-\mathbf{U}\rangle]
-\mathrm{Re}\,[\langle \mathbf{\Pi},~\mathbf{H}-\mathbf{V}\rangle]+\frac{\alpha}{2}\|\mathbf{W}-\mathbf{U}\|_F^2+\frac{\beta}{2}\|\mathbf{H}-\mathbf{V}\|_F^2
\end{array}
\end{equation}
with $\mathbf{U}\in \mathbb{Q}_\dag^{m\times l}$ and $\mathbf{V}\in \mathbb{Q}_\dag^{l\times n}$,
where $\mathbf{\Lambda}\in  \mathbb{Q}^{m\times l}$ and $\mathbf{\Pi}\in  \mathbb{Q}^{l\times n}$ are Lagrangian multipliers and $\alpha, \beta > 0$ are penalty parameters.

ADMM for (\ref{eqe1}) is given as follows
\begin{equation}\label{eqe3}
\left\{
  \begin{array}{l}
\mathbf{W}_{r+1}\leftarrow\mathrm{argmin}_{\mathbf{W}\in \mathbb{Q}^{m\times l}}  \mathcal{L}_A(\mathbf{W},\mathbf{H}_r,\mathbf{U}_r,\mathbf{V}_r,\mathbf{\Lambda}_r,\mathbf{\Pi}_r),\\
\mathbf{H}_{r+1}\leftarrow\mathrm{argmin}_{\mathbf{H}\in \mathbb{Q}^{l\times n}} \mathcal{L}_A(\mathbf{W}_{r+1},\mathbf{H},\mathbf{U}_r,\mathbf{V}_r,\mathbf{\Lambda}_r,\mathbf{\Pi}_r), \\
\mathbf{U}_{r+1}\leftarrow\mathrm{argmin}_{\mathbf{U}\in \mathbb{Q}_\dag^{m\times l}} \mathcal{L}_A(\mathbf{W}_{r+1},\mathbf{H}_{r+1},\mathbf{U},\mathbf{V}_r,\mathbf{\Lambda}_r,\mathbf{\Pi}_r), \\
\mathbf{V}_{r+1}\leftarrow\mathrm{argmin}_{\mathbf{V}\in \mathbb{Q}_\dag^{l\times n}} \mathcal{L}_A(\mathbf{W}_{r+1},\mathbf{H}_{r+1},\mathbf{U}_{r+1},\mathbf{V},\mathbf{\Lambda}_r,\mathbf{\Pi}_r), \\
\mathbf{\Lambda}_{r+1}\leftarrow\mathbf{\Lambda}_{r}-\alpha(\mathbf{W}_{r+1}-\mathbf{U}_{r+1}),  \\
\mathbf{\Pi}_{r+1}\leftarrow\mathbf{\Pi}_{r}-\beta(\mathbf{H}_{r+1}-\mathbf{V}_{r+1}).
  \end{array}
\right.
\end{equation}

Note that the gradients with respect to each quaternion variable of the augmented Lagrangian function $\mathcal{L}_A$ are
\begin{align*}
\nabla_{\mathbf{W}}\mathcal{L}_A=&(\mathbf{X}-\mathbf{W}\mathbf{H})(-\mathbf{H}^*)-\mathbf{\Lambda}+\alpha(\mathbf{W}-\mathbf{U})\\
=&\mathbf{W}(\mathbf{H}\mathbf{H}^*+\alpha I)-(\mathbf{X}\mathbf{H}^*+\mathbf{\Lambda}+\alpha \mathbf{U});\\
\nabla_{\mathbf{H}}\mathcal{L}_A=&(-\mathbf{W}^*)(\mathbf{X}-\mathbf{W}\mathbf{H})-\mathbf{\Pi}+\beta(\mathbf{H}-\mathbf{V})\\
=&(\mathbf{W}^*\mathbf{W}+\beta I)\mathbf{H}-(\mathbf{W}^*\mathbf{X}+\mathbf{\Pi}+\beta \mathbf{V});\\
\nabla_{\mathbf{U}}\mathcal{L}_A=&\mathbf{\Lambda}-\alpha(\mathbf{W}-\mathbf{U})=\alpha \mathbf{U}-(\alpha \mathbf{W}-\mathbf{\Lambda});\\
\nabla_{\mathbf{V}}\mathcal{L}_A=&\mathbf{\Pi}-\beta(\mathbf{H}-\mathbf{V})=\beta \mathbf{V}-(\beta \mathbf{H}-\mathbf{\Pi}).
\end{align*}
To efficiently implement (\ref{eqe3}),
according to the first-order 
optimality analysis for the quaternion matrix optimization  problems \cite{Qi2022},
we outline the procedure for solving those subproblems (\ref{eqe3}), called the
quaternion alternating direction method of multipliers, denoted by QADMM.

%\noindent\rule[-10pt]{13 cm}{0.1em}
\begin{algorithm}{\bf Algorithm 3~~QADMM for QNQMF (\ref{eqb1})}
\label{alg5}~

\vspace{-6mm}
\noindent\rule[-10pt]{13cm}{0.05em}

{Step 0.} Given $\mathbf{X}\in \mathbb{Q}_\dag^{m\times n}$ and $\alpha,\beta>0$. Initialize any feasible
$\mathbf{W}_0\in \mathbb{Q}^{m\times l}$, $\mathbf{H}_0\in \mathbb{Q}^{l\times n}$,
$\mathbf{U}_0\in \mathbb{Q}_\dag^{m\times l}$, $\mathbf{V}_0\in \mathbb{Q}_\dag^{l\times n}$,
$\mathbf{\Lambda}_0\in \mathbb{Q}^{m\times l}$, $\mathbf{\Pi}_0\in \mathbb{Q}^{l\times n}$.
Set $r:=0$.

{Step 1.} ~Update
\begin{equation}\label{eqe4}
\left\{
  \begin{array}{l}
\mathbf{W}_{r+1}=(\mathbf{X}\mathbf{H}_r^*+\mathbf{\Lambda}_r+\alpha \mathbf{U}_r)(\mathbf{H}_r\mathbf{H}_r^*+\alpha I)^{-1},\\
\mathbf{H}_{r+1}=(\mathbf{W}_{r+1}^*\mathbf{W}_{r+1}+\beta I)^{-1}(\mathbf{W}_{r+1}\mathbf{X}+\mathbf{\Pi}_r+\beta \mathbf{V}_r), \\
\mathbf{U}_{r+1}=\mathcal{P}_{\mathbb{Q}_\dag^{m\times l}}(\mathbf{W}_{r+1}-\frac{1}{\alpha}\mathbf{\Lambda}_r), \\
\mathbf{V}_{r+1}=\mathcal{P}_{\mathbb{Q}_\dag^{l\times n}}(\mathbf{H}_{r+1}-\frac{1}{\beta}\mathbf{\Pi}_r), \\
\mathbf{\Lambda}_{r+1}=\mathbf{\Lambda}_{r}-\alpha(\mathbf{W}_{r+1}-\mathbf{U}_{r+1}),  \\
\mathbf{\Pi}_{r+1}=\mathbf{\Pi}_{r}-\beta(\mathbf{H}_{r+1}-\mathbf{V}_{r+1}). \\
  \end{array}
\right.
\end{equation}

{Step 2.} If the stop termination criteria is satisfied, break; else, set $r:=r+1$, go to {Step 1}.

\end{algorithm}
%
%\vspace{-6mm}
%\noindent\rule[-10pt]{13cm}{0.05em}

\medskip

Now, we provide a preliminary convergent property of the proposed QADMM.

\begin{theorem}\label{th5.1}
Let $\{(\mathbf{W}_r,\mathbf{H}_r,\mathbf{U}_r,\mathbf{V}_r,\mathbf{\Lambda}_r,\mathbf{\Pi}_r)\}_{r=0}^{+\infty}$ be the sequence generated by Algorithm 3.
If
$$
\lim_{r\rightarrow +\infty}(\mathbf{W}_r,\mathbf{H}_r,\mathbf{U}_r,\mathbf{V}_r,\mathbf{\Lambda}_r,\mathbf{\Pi}_r)=(\widehat{\mathbf{W}},\widehat{\mathbf{H}},\widehat{\mathbf{U}},\widehat{\mathbf{V}},\widehat{\mathbf{\Lambda}},\widehat{\mathbf{\Pi}})
,$$
then $(\widehat{\mathbf{W}},\widehat{\mathbf{H}})$ is a stationary point of QNQMF (\ref{eqb4}).
\end{theorem}

\textbf{\emph{Proof}} Firstly, we prove that
\begin{equation}\label{eqe5}
\left\{
  \begin{array}{llllll}
\mathbf{U}_r\in  \mathbb{Q}_\dag^{m\times l},&
\mathbf{\Lambda}_r\in  \mathbb{Q}_\dag^{m\times l},& \mathrm{Re}\,\mathbf{\Lambda}_r=0,\\
\mathrm{Im}_i\,\mathbf{U}_r \odot \mathrm{Im}_i\,\mathbf{\Lambda}_r=0, &
\mathrm{Im}_j\,\mathbf{U}_r \odot \mathrm{Im}_j\,\mathbf{\Lambda}_r=0, &
\mathrm{Im}_k\,\mathbf{U}_r \odot \mathrm{Im}_k\,\mathbf{\Lambda}_r=0,
\\
\mathbf{V}_r\in  \mathbb{Q}_\dag^{l\times n},&
\mathbf{\Pi}_r\in  \mathbb{Q}_\dag^{l\times n},&
\mathrm{Re}\,\mathbf{\Pi}_r=0, \\
\mathrm{Im}_i\,\mathbf{V}_r \odot \mathrm{Im}_i\,\mathbf{\Pi}_r=0, &
\mathrm{Im}_j\,\mathbf{V}_r \odot \mathrm{Im}_j\,\mathbf{\Pi}_r=0, &
\mathrm{Im}_k\,\mathbf{V}_r \odot \mathrm{Im}_k\,\mathbf{\Pi}_r=0 \\
  \end{array}
\right.
\end{equation}
for $r=0,1,2,\ldots$.
In fact, denote $\mathbf{D}_r=\mathbf{W}_{r+1}-\frac{1}{\alpha}\mathbf{\Lambda}_r=D_0+D_1 \mathbf{i}+D_2 \mathbf{j}+D_3 \mathbf{k}$, then
$$\mathbf{U}_{r+1}=D_0+\max(D_1,0)\mathbf{i}+\max(D_2,0)\mathbf{j}+\max(D_3,0)\mathbf{k}\in \mathbb{Q}_\dag^{m\times l}.$$
Note that
\begin{eqnarray*}
&&\frac{1}{\alpha}\mathbf{\Lambda}_{r+1}=(\frac{1}{\alpha}\mathbf{\Lambda}_r-\mathbf{W}_{r+1})+\mathbf{U}_{r+1}=-\mathbf{D}_r+\mathbf{U}_{r+1}\\
&&=-D_0-D_1 \mathbf{i}-D_2 \mathbf{j}-D_3 \mathbf{k}+D_0+\max(D_1,0)\mathbf{i}+\max(D_2,0)\mathbf{j}+\max(D_3,0)\mathbf{k}\\
&&=(\max(D_1,0)-D_1)\mathbf{i}+(\max(D_2,0)-D_2)\mathbf{j}+(\max(D_3,0)-D_3)\mathbf{k}.
\end{eqnarray*}
Since $$\max(a,0)-a=\frac{1}{2}(|a|+a)-a=\frac{1}{2}(|a|-a)\geq 0$$ for any real number $a$,
it has
$$\max(D_1,0)-D_1\geq 0,\quad \max(D_2,0)-D_2\geq 0,\quad \max(D_3,0)-D_3\geq 0.$$
Thus, $\mathbf{\Lambda}_{r+1}\in \mathbb{Q}_\dag^{m\times l}$ as $\alpha>0$ and $\mathrm{Re}\,\mathbf{\Lambda}_{r+1}=0$.
Again, since $$\max(a,0)[\max(a,0)-a]=\frac{1}{4}(|a|+a)(|a|-a)=0$$  for any real number $a$, it follows that
$$\mathrm{Im}_i\,\mathbf{U}_r \odot \mathrm{Im}_i\,\mathbf{\Lambda}_r=0,\quad
\mathrm{Im}_j\,\mathbf{U}_r \odot \mathrm{Im}_j\,\mathbf{\Lambda}_r=0,\quad
\mathrm{Im}_k\,\mathbf{U}_r \odot \mathrm{Im}_k\,\mathbf{\Lambda}_r=0.$$
Hence, the first relations of (\ref{eqe5}) hold.
Similarly, the second relations of (\ref{eqe5}) also hold.

Next, taking the limit in (\ref{eqe4}) as $r\rightarrow +\infty$, we can obtain
\begin{eqnarray*}
\left\{
  \begin{array}{l}
\widehat{\mathbf{W}}(\widehat{\mathbf{H}}\widehat{\mathbf{H}}^*+\alpha I)=\mathbf{X}\widehat{\mathbf{H}}^*+\widehat{\mathbf{\Lambda}}+\alpha \widehat{\mathbf{U}},\\
(\widehat{\mathbf{W}}^*\widehat{\mathbf{W}}+\beta I)\widehat{\mathbf{H}}=\widehat{\mathbf{W}}^*\mathbf{X}+\widehat{\mathbf{\Pi}}+\beta \widehat{\mathbf{V}}, \\
\widehat{\mathbf{\Lambda}}=\widehat{\mathbf{\Lambda}}-\alpha(\widehat{\mathbf{W}}-\widehat{\mathbf{U}}),  \\
\widehat{\mathbf{\Pi}}=\widehat{\mathbf{\Pi}}-\beta(\widehat{\mathbf{H}}-\widehat{\mathbf{V}}). \\
  \end{array}
\right.
\end{eqnarray*}
Hence, we have
\begin{equation}\label{eqe6}
\left\{
  \begin{array}{l}
\widehat{\mathbf{U}}=\widehat{\mathbf{W}},\quad \widehat{\mathbf{\Lambda}}=-(\mathbf{X}-\widehat{\mathbf{W}}\widehat{\mathbf{H}})\widehat{\mathbf{H}}^*,\\
\widehat{\mathbf{V}}=\widehat{\mathbf{H}},\quad \widehat{\mathbf{\Pi}}=-\widehat{\mathbf{W}}^*(\mathbf{X}-\widehat{\mathbf{W}}\widehat{\mathbf{H}}).
  \end{array}
\right.
\end{equation}
From (\ref{eqe5}) and (\ref{eqe6}), it can get that $(\widehat{\mathbf{W}},\widehat{\mathbf{H}})$ satisfies (\ref{eqb6}), that is a stationary point of QNQMF (\ref{eqb4}).
This completes the proof.
\hfill\endproof

\medskip

\begin{remark}
For the NMF problem (\ref{eqa1}), the process  (\ref{eqe4}) reduces to
\begin{equation}\label{eqe8}
\left\{
  \begin{array}{l}
W_{r+1}=(XH_r^T+\Lambda_r+\alpha U_r)(H_rH_r^T+\alpha I)^{-1},\\
H_{r+1}=(W_{r+1}^TW_{r+1}+\beta I)^{-1}(W_{r+1}X+\Pi_r+\beta V_r), \\
U_{r+1}=\max(W_{r+1}-\frac{1}{\alpha}\Lambda_r,~0), \\
V_{r+1}=\max(H_{r+1}-\frac{1}{\beta}\Pi_r,~0), \\
\Lambda_{r+1}=\Lambda_{r}-\alpha(W_{r+1}-U_{r+1}),  \\
\Pi_{r+1}=\Pi_{r}-\beta(H_{r+1}-V_{r+1}), \\
  \end{array}
\right.
\end{equation}
which is the algorithm proposed in \cite{NMF_ADMM1}.
\end{remark}

\section{Applications and numerical experiments}\label{sec5}

In this section, we utilize some test problems to examine the effectiveness of the proposed algorithms.
All test problems are performed under
MATLAB R2020b on a personal computer with 3.50 GHz central processing unit (Gen Intel(R) Core(TM)  i9-11900K), 32.00 GB
memory and Windows 10 operating system.

\subsection{Color image reconstruction}\label{sec5.1}
In this subsection, we focus QNQMF on the color image  reconstruction.
In order to evaluate the quality of  reconstructed images, we employ the peak signal
to noise ratio (PSNR) criterion. The PSNR of $m\times n$ color images $\mathbf{X}=X_R{\bf i} + X_G{\bf j} + X_B{\bf k}$ and $\mathbf{Z}=Z_R {\bf i} + Z_G {\bf j} + Z_B {\bf k}$ is calculated as follows:
\begin{eqnarray*}
\mathrm{PSNR}=20\,\mathrm{log}_{10}\frac{255}{\mathrm{MSE}}~~[\mathrm{dB}],\quad
\mathrm{MSE}=\sqrt{{\frac{1}{mn}\sum_{s=1}^m\sum_{t=1}^n |\mathbf{X}_{st}-\mathbf{Z}_{st}|^2}},
\end{eqnarray*}
where MSE is called the mean square error.
In addition,  `Time' denotes the elapsed CPU time in seconds.

When the red, green and blue color channels of the original color image are regarded as the image non-negative matrices,  denoted by $X_R$, $X_G$ and $X_B$, and
then the NMF can be decomposed on the image matrices $X_R$, $X_G$ and $X_B$, that is
\begin{equation}\label{eqke1}
X_R=W_RH_R,\quad  X_G=W_GH_G,\quad X_B=W_BH_B ,
\end{equation}
where $W_R,W_G,W_B\in \mathbb{R}^{m\times l}$ and $H_R,H_G,H_B\in \mathbb{R}^{l\times n}$ are the non-negative matrices.
Let $r$ be the iterations. Denote
$$\mathbf{Z}_r=\mathrm{Im}\,(\mathbf{W}_{r}\mathbf{H}_{r}),\quad \mathrm{RES}(r)=\|\mathrm{Im}\,\mathbf{X}-\mathbf{Z}_r\|_F$$
for QNQMF and
$$\mathbf{Z}_r=(W_R)_{r}(H_R)_{r}{\bf i}+(W_G)_{r}(H_G)_{r}{\bf j}+(W_B)_{r}(H_B)_{r}{\bf k},\quad \mathrm{RES}(r)=\|\mathrm{Im}\,\mathbf{X}-\mathbf{Z}_r\|_F$$
for NMF  (\ref{eqke1}).

\medskip

Here, we compare the following four methods.
\begin{itemize}
  \item QIPGM:  solve QNQMF (\ref{eqb1}) by Algorithm 2 with $\rho=0.01$ and  $\sigma=0.001$;
  \item QADMM:  solve QNQMF (\ref{eqb1}) by Algorithm 3 with $\alpha=0.01$ and $\beta=0.01$;
  \item RIPGM:  solve NMF (\ref{eqke1}) by \cite{NMF_Lin} with $\rho=0.01$ and  $\sigma=0.001$;
  \item RADMM:  solve NMF (\ref{eqke1}) by (\ref{eqe8}) with $\alpha=0.01$ and $\beta=0.01$.
\end{itemize}

The choices of these numerical parameters are based on the experiments and the same values
are used for the quaternion and real algorithms.
Let
$$
  \begin{array}{lll}
L_1=\mathrm{rand}(m,l),& L_2=\mathrm{rand}(m,l), & L_3=\mathrm{rand}(m,l),\\
S_1=\mathrm{rand}(l,n),& S_2=\mathrm{rand}(l,n), & S_3=\mathrm{rand}(l,n).
  \end{array}
$$
The choices of the initial matrices for different algorithms are presented in Table \ref{tab0}.

\begin{table}[!ht]
\caption{The choice of the initial matrices.}
\vspace{-4mm}
\begin{center} \footnotesize
\begin{tabular}{|c|l|} \hline
Method & Initial matrices \\\hline
QIPGM   &
$
\begin{array}{l}
\mathbf{W}_0=L_1 \mathbf{i}+L_2 \mathbf{j}+L_3 \mathbf{k},\\
\mathbf{H}_0=S_1 \mathbf{i}+S_2 \mathbf{j}+S_3 \mathbf{k}\\
\end{array}
$\\\hline

QADMM   &
$
\begin{array}{l}
\mathbf{W}_0=\mathbf{U}_0=\mathbf{\Lambda}_0=L_1 \mathbf{i}+L_2 \mathbf{j}+L_3 \mathbf{k},\\
\mathbf{H}_0=\mathbf{V}_0=\mathbf{\Pi}_0=S_1 \mathbf{i}+S_2 \mathbf{j}+S_3 \mathbf{k}\\
\end{array}
$\\\hline

RIPGM   &
$
  \begin{array}{lll}
\text{Red color channel:}   & (W_R)_0=L_1,  & (H_R)_0=S_1\\
\text{Green color channel:} & (W_G)_0=L_2,  & (H_G)_0=S_2\\
\text{Blue color channel:}  & (W_B)_0=L_3,  & (H_B)_0=S_3\\
  \end{array}
$\\\hline

RADMM   &
$
  \begin{array}{lll}
\text{Red color channel:}   & (W_R)_0=(U_R)_0=(\Lambda_R)_0=L_1,  & (H_R)_0=(V_R)_0=(\Pi_R)_0=S_1\\
\text{Green color channel:} & (W_G)_0=(U_G)_0=(\Lambda_G)_0=L_2,  & (H_G)_0=(V_G)_0=(\Pi_G)_0=S_2\\
\text{Blue color channel:}  & (W_B)_0=(U_B)_0=(\Lambda_B)_0=L_3,  & (H_B)_0=(V_B)_0=(\Pi_B)_0=S_3\\
  \end{array}
$\\\hline
\end{tabular}
\end{center}\label{tab0}
\end{table}

\begin{figure}[!ht]
 \begin{center}
\includegraphics[height=0.30\linewidth,width=0.8\linewidth]{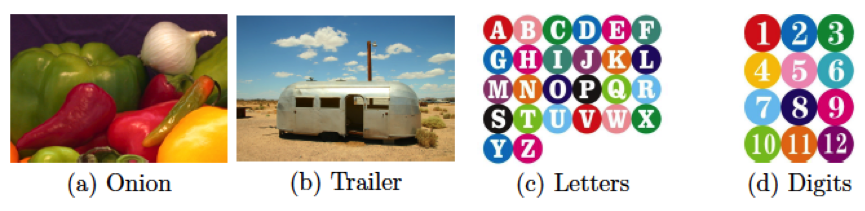}
\end{center}
  \caption{Experimental color images.}\label{Fig1}
\end{figure}

\begin{figure}[!ht]
\begin{center}
\includegraphics[height=0.45\linewidth,width=0.8\linewidth]{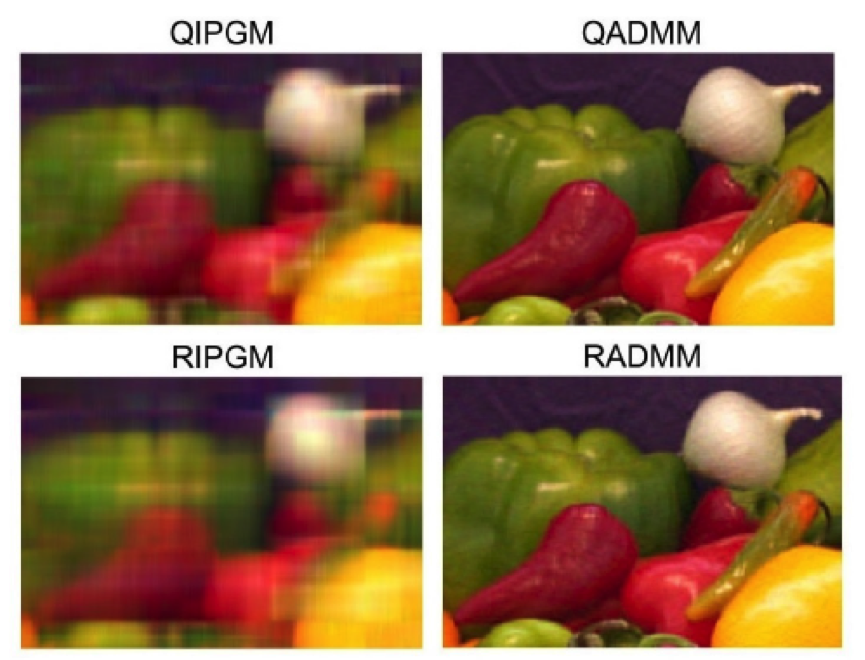}
\end{center}
\caption{Experimental results of four different algorithms for Onion with $l=40$.}\label{Fig2}
\end{figure}

\begin{figure}[!ht]
\begin{center}
\includegraphics[height=0.45\linewidth,width=0.8\linewidth]{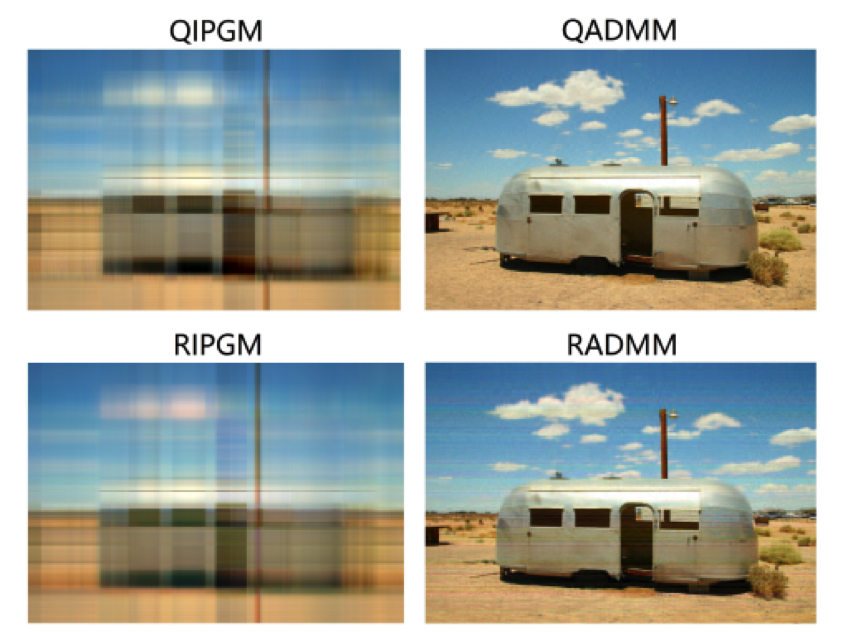}
\end{center}
\caption{Experimental results of four different algorithms for Trailer with $l=80$.}\label{Fig3}
\end{figure}

\begin{figure}[!ht]
\begin{center}
\includegraphics[height=0.45\linewidth,width=0.8\linewidth]{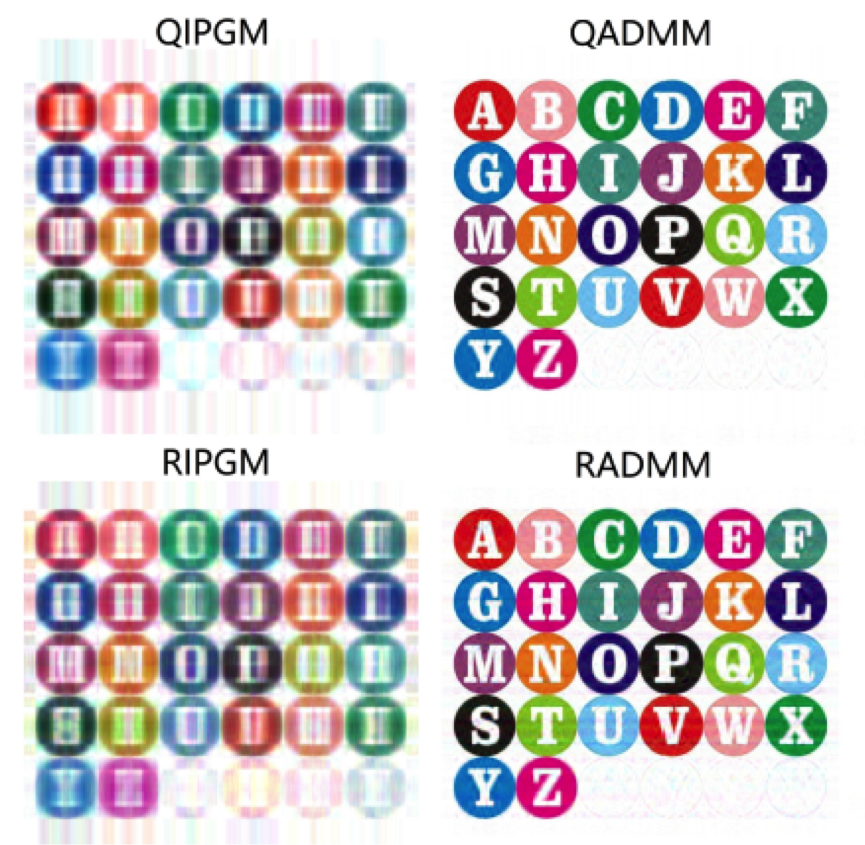}
\end{center}
 \caption{Experimental results of four different algorithms for Letters with $l=40$.}\label{Fig4}
\end{figure}

\begin{figure}[!ht]
\begin{center}
\includegraphics[height=0.45\linewidth,width=0.8\linewidth]{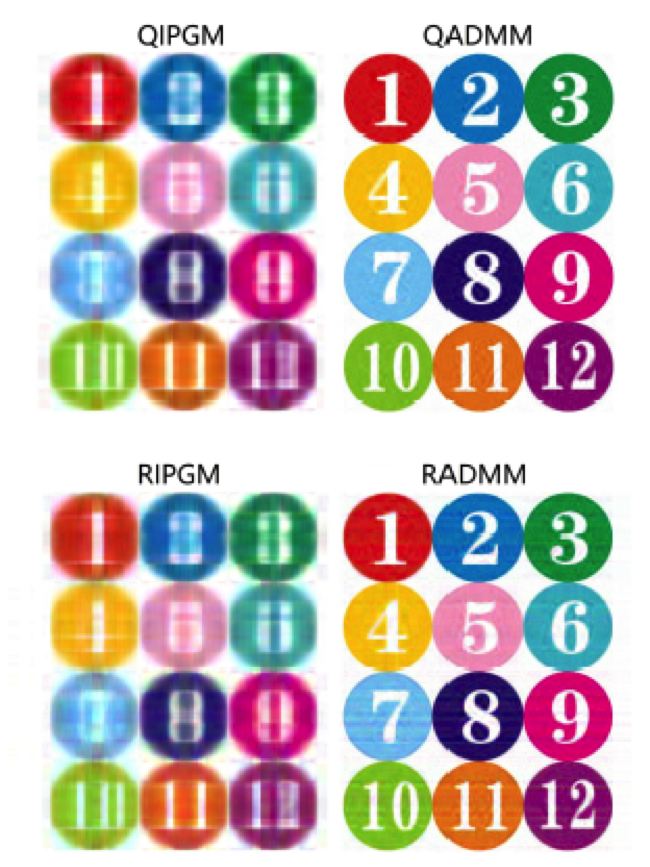}
\end{center}
 \caption{Experimental results of four different algorithms for Digits with $l=40$.}\label{Fig5}
\end{figure}

\begin{figure}[!ht]
\begin{center}
\includegraphics[height=0.85\linewidth,width=1.0\linewidth]{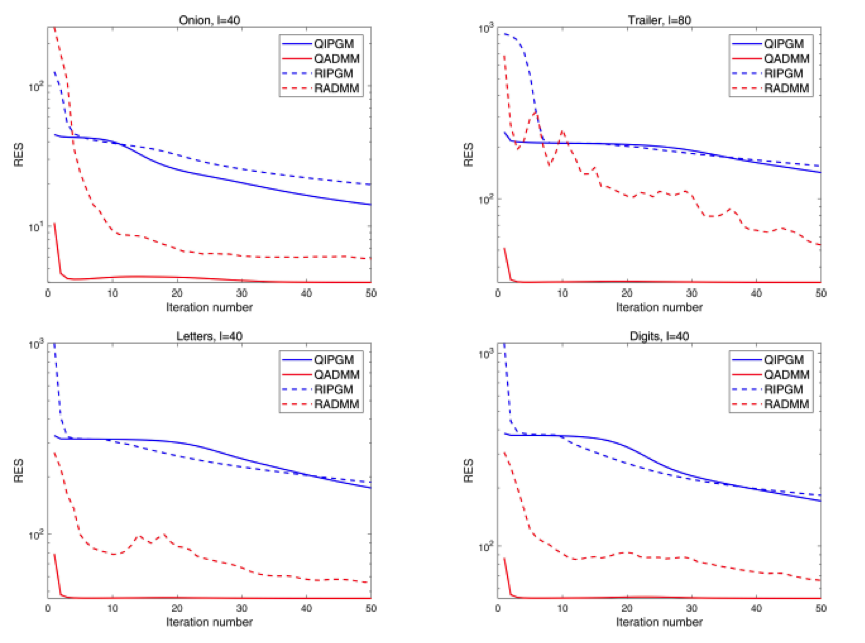}
\end{center}
  \caption{Relationship between the number of iterations and $\mathrm{log}_{10}(\mathrm{RES}(r))$ for  different color images.}\label{Fig6}
\end{figure}

We test four different color images, named  Onion,  Trailer,  Letters and  Digits; see Fig.~\ref{Fig1}.
For these color images, all algorithms run $50$ iterations and the obtained numerical results are presented on Tables \ref{tab1}-\ref{tab4}, respectively.
Figures \ref{Fig2}-\ref{Fig5} present the reconstructed color images $\mathbf{Z}_{50}$.
And the relationship between the number of iterations and $\mathrm{log}_{10}(\mathrm{RES}(r))$ for the QNQMF and
RGB three channels NMF are plot in Fig.~\ref{Fig6}.

\begin{table}[!ht]
\caption{Numerical results of the color image Onion $($$135\times 198$ pixels$)$.}
\vspace{-2mm}
\begin{center} \footnotesize
\begin{tabular}{|c|c|c|c|c|c|c|c|} \hline
$l$    & Method  & QIPGM   &  QADMM       & RIPGM  &  RADMM   \\\hline
10     & Time(s) & 0.2013  & 0.0963       & 0.0146 & 0.0064\\
       & PSNR    & 79.8477 & \bf{81.4198} & 79.2830& 80.4876\\ \hline

20     & Time(s) & 0.2456  & 0.1434       & 0.0160 & 0.0076\\
       & PSNR    & 80.4340 & \bf{83.4851} & 79.4373& 81.6233\\ \hline

30     & Time(s) & 0.2953  & 0.1847       & 0.0211 & 0.0100\\
       & PSNR    & 80.7172 & \bf{85.0360} & 79.5825& 83.9558\\ \hline

40     & Time(s) & 0.3398  & 0.2243       & 0.0214 & 0.0135\\
       & PSNR    & 80.8676 & \bf{86.4066} & 79.4339& 84.7264\\ \hline
\end{tabular}
\end{center}\label{tab1}
\end{table}

\begin{table}[!ht]
\caption{Numerical results of the color image Trailer $($$683\times 1024$ pixels$)$.}
\vspace{-2mm}
\begin{center} \footnotesize
\begin{tabular}{|c|c|c|c|c|c|c|c|} \hline
$l$    & Method  & QIPGM   &  QADMM       & RIPGM  &  RADMM   \\\hline
20     & Time(s) & 9.7480  & 3.2211       & 0.8885 & 0.1091\\
       & PSNR    & 84.7805 & \bf{88.5377} & 85.8208& 88.2529\\ \hline

40     & Time(s) & 10.2728 & 3.6850       & 0.9614 & 0.1445\\
       & PSNR    & 84.9515 & \bf{89.8671} & 86.2576& 89.0544\\ \hline

60     & Time(s) & 11.3930 & 4.2889       & 1.0765 & 0.1748\\
       & PSNR    & 85.0318 & \bf{90.7518} & 86.1012& 89.0008\\ \hline

80     & Time(s) & 12.0315 & 4.7420       & 0.8466 & 0.2031\\
       & PSNR    & 85.0520 & \bf{91.4671} & 84.6746& 89.2806\\ \hline
\end{tabular}
\end{center}\label{tab2}
\end{table}

\begin{table}[!ht]
\caption{Numerical results of the color image Letters  $($$680\times 680$ pixels$)$.}
\vspace{-2mm}
\begin{center} \footnotesize
\begin{tabular}{|c|c|c|c|c|c|c|c|} \hline
$l$    & Method  & QIPGM   &  QADMM       & RIPGM  &  RADMM   \\\hline
10     & Time(s) & 5.8279  & 2.0546       & 0.5159 & 0.0618\\
       & PSNR    & 81.7718 & \bf{83.6386} & 81.8125& 83.5888\\ \hline

20     & Time(s) & 6.4522  & 2.1344       & 0.5311 & 0.0936\\
       & PSNR    & 82.1142 & \bf{85.5912} & 81.8661& 85.3397\\ \hline

30     & Time(s) & 6.4122  & 2.1437       & 0.5181 & 0.0744\\
       & PSNR    & 82.2917 & \bf{87.0045} & 82.1053& 86.4040\\ \hline

40     & Time(s) & 6.5475  & 2.2501       & 0.5641 & 0.0911\\
       & PSNR    & 82.3586 & \bf{88.1478} & 82.0681& 87.2978\\ \hline
\end{tabular}
\end{center}\label{tab3}
\end{table}

\begin{table}[!ht]
\caption{Numerical results of the color image Digits  $($$800\times 800$ pixels$)$.}
\vspace{-2mm}
\begin{center} \footnotesize
\begin{tabular}{|c|c|c|c|c|c|c|c|} \hline
$l$    & Method  & QIPGM   &  QADMM       & RIPGM  &  RADMM   \\\hline
10     & Time(s) & 8.5460  & 2.5299       & 0.8968 & 0.0930\\
       & PSNR    & 83.2662 & \bf{85.0228} & 82.9325& 84.9425\\ \hline

20     & Time(s) & 8.7719  & 2.5244       & 0.8720 & 0.0989\\
       & PSNR    & 83.6544 & \bf{86.7386} & 83.2849& 86.3813\\ \hline

30     & Time(s) & 8.9382  & 2.7024       & 0.8939 & 0.1104\\
       & PSNR    & 83.8515 & \bf{87.9248} & 83.4227& 87.2967\\ \hline

40     & Time(s) & 9.2815  & 2.8476       & 0.9283 & 0.1337\\
       & PSNR    & 83.8593 & \bf{88.9088} & 83.5650& 87.9792\\ \hline
\end{tabular}
\end{center}\label{tab4}
\end{table}

From these numerical results, the following facts can be observed.
\begin{itemize}
\item  QIPGM and QADMM cost more CPU time than  RIPGM and RADMM  at one iteration step. However, the PSNR values of  QIPGM and QADMM are better than  RIPGM and RADMM, respectively.
    In other words,  these algorithms based on QNQMF take more CPU time at one iteration step, but they have better numerical performances than the corresponding algorithms based on NMF.

\item  For QIPGM and RIPGM, the sequences $\{\mathrm{RES}(r)\}$ are monotonically nonincreasing from
    Fig.~\ref{Fig6}, which is a manifestation of Lemma \ref{lem4.4}.
    Since QIPGM and RIPGM need to search the step sizes $\alpha_r$ and $\beta_r$ at each iteration step,  it will take a lot of CPU  time.  Thus, the CPU time of QIPGM is the maximum of these compared algorithms.

\item The PSNR value  of QADMM is the best from Tables \ref{tab1}-\ref{tab4}. In addition, we can see that QADMM just need few number of iterations  to reach the minimum residual value $\mathrm{RES}(r)$; see Fig.~\ref{Fig6}.
    Moreover, the recovered image via QADMM is the sharpest in the image quality; see
    Figures \ref{Fig2}-\ref{Fig5}.
In particular, for QADMM and RADMM,  it is very obvious  increase in the PSNR value as the value $l$ increases.
\end{itemize}

\subsection{Color face recognition}\label{sec5.2}
Suppose that $\mathbf{Q}=Q_R{\bf i} + Q_G{\bf j} + Q_B{\bf k}\in \mathbb{Q}^{m\times n}$ represents a color face image with $Q_R$, $Q_G$ and $Q_B$ being respectively the red, green, and blue values of pixels.
Suppose that the training set contains $\mu$ known face image, denoted by
$\mathbb{T}=\{\mathbf{F}_1,\mathbf{F}_2,\ldots,\mathbf{F}_{\mu}\}$ and the testing set contains $\nu$ face image to be recognized,
denoted by $\mathbb{S}=\{\mathbf{G}_1,\mathbf{G}_2,\ldots,\mathbf{G}_{\nu}\}$. Here, $\mathbf{F}_{t}(t=1,2,\ldots,\mu)$ and $\mathbf{G}_{s}(s=1,2,\ldots,\nu)$ are $mn\times 1$ pure quaternion vectors,
which are obtained via straighten $m\times n$ color face images into the columns.

Let $\mathbf{X}=(\mathbf{F}_1,\mathbf{F}_2,\ldots,\mathbf{F}_{\mu})$ and
$l$ be the positive integer. Assume that
the basis images are $\mathbf{W}=(\mathbf{w}_1,\mathbf{w}_2,\ldots,\mathbf{w}_l)$ and the encodings are $\mathbf{H}=(\mathbf{h}_1,\mathbf{h}_2,\ldots,\mathbf{h}_{\mu})$.
Since the QNQMF
$$\mathbf{X}=\mathbf{W}\mathbf{H},$$ the each face $\mathbf{F}_t$  can be approximately
reconstructed by
$$\mathbf{F}_t=\mathbf{W}\mathbf{h}_t,\quad t=1,2,\ldots,\mu.$$

In Algorithm \ref{alg6}, we present the color face  recognition based on the QNQMF.

\noindent\rule[-10pt]{13 cm}{0.1em}
\begin{algorithm}%[Color face  recognition based on QNQMF]
\label{alg6}~

\vspace{-6mm}
\noindent\rule[-10pt]{13cm}{0.05em}

{Step 0.} Input the color face images in
$$\mathbb{T}=\{\mathbf{F}_1,\mathbf{F}_2,\ldots,\mathbf{F}_{\mu}\}\quad and \quad \mathbb{S}=\{\mathbf{G}_1,\mathbf{G}_2,\ldots,\mathbf{G}_{\nu}\}.$$
Let the  training data $\mathbf{X}_{train}=(\mathbf{F}_1,\mathbf{F}_2,\ldots,\mathbf{F}_{\mu})$.

{Step 1.} Do QNQMF:
$$\mathbf{X}_{train}=\mathbf{W}_{train}\mathbf{H}_{train}.$$

{Step 2.} For  $s=1,2,\ldots,\nu$, compute
$$\mathbf{h}_s=(\mathbf{W}_{train}^*\mathbf{W}_{train})^{-1}(\mathbf{W}_{train}^*\mathbf{G}_s) $$
and
\begin{equation}\label{eqke3}
\theta_{t,s}=\frac{\mathrm{Re}\,\langle \mathbf{h}_s,~\mathbf{H}_{train}(:,t)\rangle}{\|\mathbf{h}_s\|_F\|\mathbf{H}_{train}(:,t)\|_F },\quad t=1,\ldots,\mu.
\end{equation}
  Let
$$t_{\circledast}=\mathrm{argmax}_{1\leq t\leq \mu}\, \theta_{t,s}.$$
Such $\mathbf{F}_{t_{\circledast}}$ is the optimum matching encoding for the trained image $\mathbf{G}_s$.
\end{algorithm}

\vspace{-6mm}
\noindent\rule[-10pt]{13cm}{0.05em}

\medskip

In (\ref{eqke3}), $\mathbf{H}_{train}(:,t)$ denotes the $t$-th column of the  quasi non-negative quaternion  matrix $\mathbf{H}_{train}$
and the  similarity measure $\theta_{t,s}$ determines the matching score between  $\mathbf{h}_s$ and $\mathbf{H}_{train}(:,t)$, which is  corresponding to $\mathbf{G}_s$ in  the testing set and $\mathbf{F}_t$ in  the training set. When we test on the gray face images,
${h}_s$ and ${H}_{train}(:,t)$ reduce to the real vectors and the value
\begin{equation}\label{eqke4}
\theta_{t,s}=\frac{\langle {h}_s,~{H}_{train}(:,t)\rangle}{\|{h}_s\|_2\|{H}_{train}(:,t)\|_2 }
\end{equation}
is the cosine angle between the two  vectors.
Hence, the  maximum similarity measure  means the best matching trained  image.
We remark that the authors in \cite{ADMM_color}  divided the color image into the red, green, blue color  channels
and proposed the following Algorithm 4.

\noindent\rule[-10pt]{13 cm}{0.1em}
\begin{algorithm}{\bf Algorithm 4~~Color face  recognition based on color channels NMF \cite{ADMM_color}}
\label{alg7}~

\vspace{-6mm}
\noindent\rule[-10pt]{13cm}{0.05em}

{Step 0.} Input the color face images in
$$\mathbb{T}=\{\mathbf{F}_1,\mathbf{F}_2,\ldots,\mathbf{F}_{\mu}\}\quad and \quad \mathbb{S}=\{\mathbf{G}_1,\mathbf{G}_2,\ldots,\mathbf{G}_{\nu}\},$$
where
$\mathbf{F}_t=F_{R,t} \mathbf{i}+F_{G,t}\mathbf{j}+F_{B,t}\mathbf{k}$ for $t=1,2,\ldots,\mu$
and
$\mathbf{G}_s=G_{R,s} \mathbf{i}+G_{G,s}\mathbf{j}+G_{B,s}\mathbf{k}$ for $s=1,2,\ldots,\nu$.
Let the  training matrices
$$
\left\{
\begin{array}{l}
X_{R,train}=(F_{R,1},F_{R,2},\ldots,F_{R,\mu}),\\
X_{G,train}=(F_{G,1},F_{G,2},\ldots,F_{G,\mu}),\\
X_{B,train}=(F_{B,1},F_{B,2},\ldots,F_{B,\mu})
\end{array}
\right.$$
be the red,
green, and blue values of pixels, respectively.

{Step 1.} Do NMF on real field:
$$
\left\{
  \begin{array}{l}
X_{R,train}=W_{R,train}H_{R,train}, \\
X_{G,train}=W_{G,train}H_{G,train}, \\
X_{B,train}=W_{B,train}H_{B,train}. \\
  \end{array}
\right.
$$

{Step 2.} For  $s=1,2,\ldots,\nu$, compute
$$
\left\{
  \begin{array}{l}
h_{R,s}=(W_{R,train}^*W_{R,train})^{-1}(W_{R,train}^*G_{R,s}),\\
h_{G,s}=(W_{G,train}^*W_{G,train})^{-1}(W_{G,train}^*G_{G,s}),\\
h_{B,s}=(W_{B,train}^*W_{B,train})^{-1}(W_{B,train}^*G_{B,s})
  \end{array}
\right.
$$
and
\begin{eqnarray*}
\theta_{t,s}=\frac{\langle h_{R,s},\,H_{R,train}(:,t)\rangle}{\|h_{R,s}\|_2\|H_{R,train}(:,t)\|_2 }+
\frac{\langle h_{G,s},\,H_{G,train}(:,t)\rangle}{\|h_{G,s}\|_2\|H_{G,train}(:,t)\|_2 }+
\frac{\langle h_{B,s},\,H_{B,train}(:,t)\rangle}{\|h_{B,s}\|_2\|H_{B,train}(:,t)\|_2 }
\end{eqnarray*}
for $t=1,\ldots,\mu$.
Let
$$t_{\circledast}=\mathrm{argmax}_{1\leq t\leq \mu}\, \theta_{t,s}.$$
Such $\mathbf{F}_{t_{\circledast}}$ is the optimum matching encoding for the trained image $\mathbf{G}_s$.
\end{algorithm}

%\vspace{-6mm}
%\noindent\rule[-10pt]{13cm}{0.05em}

\medskip

In this subsection, we utilize QNQMF to the color face recognition. From the results presented in Subsection \ref{sec5.1}, we know that the numerical performance of ADMM-like method is good. Thus, we study the ADMM-like method for the face recognition.

\medskip

Here, we compare the following these algorithms.
\begin{itemize}
\item QADMM-color: Algorithm \ref{alg6} with Algorithm 3  for the color face images;

\item RADMM-color: Algorithm 4 with the iteration (\ref{eqe8})   for the color face images;

\item RADMM-gray:  the iteration (\ref{eqe8})  with (\ref{eqke4}) for the gray face images;

\item QPCA-color: the quaternion principal component analysis for  the color face images \cite{QPCA_Lancos}.
\end{itemize}
In addition, the paraments $\alpha=0.01$ and $\beta=0.01$ for QADMM-color, RADMM-color and RADMM-gray methods.
For QPCA-color, the value $l$ means the first $l$ largest eigenvalues and corresponding eigenvectors. The  accuracy rate is defined as
$$\mathrm{Rate}=\dfrac{\text{True~Numbers}}{\text{Total Numbers}}.$$

\begin{figure}[!ht]
\begin{center}
\includegraphics[height=0.45\linewidth,width=1.0\linewidth]{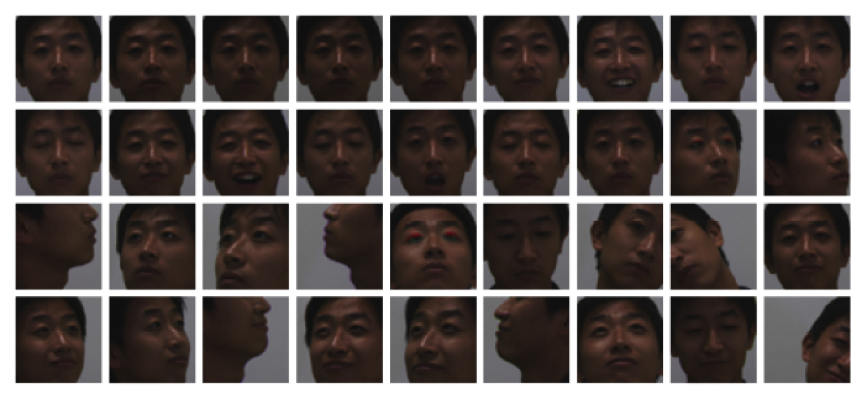}
\end{center}
\caption{ Cropped   color face images  from  CASIA 3D face database. }\label{Fig9}
\end{figure}

\begin{example}\label{ex6.2}
In this experiment, we focus on the famous  CASIA 3D face database \cite{data_CASIA}.
CASIA 3D face database  involves $123$ individuals and each individual contains $37$ or $38$ different color face images with $640\times 480$ pixels.
We focus on the first $37$ images for each individual.
The training set is randomly selected with the  random number $\eta(=20,24,28,32)$
and the remaining as the testing set.
All the color face images are cropped as the $100\times 100$ pixels, then  resized to $50\times 50$ pixels.
Fig.~\ref{Fig9} presents the faces from the first one $($just the first $36$ images$)$.
The pictures show  not only the single variations of poses, expressions and illuminations, but also the combined variations of expressions under illumination and poses under expressions.
We preserve a small amount of face information for some images after processing.
\end{example}

\medskip

\begin{table}[!ht]
\caption{Numerical results of Example \ref{ex6.2} $(\eta=20)$.}
\begin{center} \footnotesize
\begin{tabular}{|c|c|c|c|c|c|c| } \hline
Method         &$l$      &  $3$    & $6$       &  $9$     & $12$      & $15$       \\\hline
QADMM-color    & Time(s) &33.2277  &  33.6415  & 34.0056  &  34.5794  & 36.7154    \\
               & Rate    &26.88$\%$&  51.79$\%$& 67.10$\%$&  70.73$\%$& 73.17$\%$  \\\hline

RADMM-color    & Time(s) &5.5573   &  5.6143   & 5.7008   &  5.7749   & 5.9524     \\
               & Rate    &18.08$\%$&  51.89$\%$& 60.78$\%$&  68.48$\%$& 71.54$\%$  \\\hline

RADMM-gray     & Time(s) &2.8677   &  2.8651   & 2.9298   &  2.9524   & 3.0534     \\
               & Rate    &6.70$\%$ &  38.35$\%$& 58.97$\%$&  63.27$\%$& 67.53$\%$  \\\hline

QPCA-color     & Time(s) &54.1980  & 60.3608   &75.4179   & 87.1528   &90.9380     \\
               & Rate    &37.64$\%$&  58.54$\%$& 62.94$\%$&  66.04$\%$& 67.29$\%$  \\\hline
\end{tabular}
\end{center}\label{tab9}
\end{table}

%\medskip

\begin{table}[!ht]
\caption{Numerical results of Example \ref{ex6.2} $(\eta=24)$.}
\begin{center} \footnotesize
\begin{tabular}{|c|c|c|c|c|c|c| } \hline
Method         &$l$      &  $3$    & $6$        &  $9$        & $12$       & $15$        \\\hline
QADMM-color    & Time(s) &56.4495  &  59.3814   &  45.3469    &  32.6791   &  58.4169    \\
               & Rate    &24.77$\%$&   45.65$\%$&   63.85$\%$ &   68.04$\%$&   72.23$\%$ \\\hline

RADMM-color    & Time(s) &8.9977   &   9.3838   &   5.2911    &   5.3229   &   8.9831    \\
               & Rate    &16.76$\%$&   45.22$\%$&   58.79$\%$ &   68.04$\%$&   70.04$\%$ \\\hline

RADMM-gray     & Time(s) &3.4718   &   3.6892   &   2.6669    &   2.6659   &   3.6898    \\
               & Rate    &6.32$\%$ &   34.83$\%$&   54.91$\%$ &   64.23$\%$&   67.85$\%$ \\\hline

QPCA-color     & Time(s) &72.2730  & 100.6967   & 117.7804    &  84.2996   &  99.1015    \\
               & Rate    &38.34$\%$&   57.22$\%$&   62.85$\%$ &   64.60$\%$&   67.92$\%$ \\\hline
\end{tabular}
\end{center}\label{tab10}
\end{table}

\begin{table}[!ht]
\caption{Numerical results of Example \ref{ex6.2} $(\eta=28)$.}
\begin{center} \footnotesize
\begin{tabular}{|c|c|c|c|c|c|c| } \hline
Method         &$l$      &  $3$    & $6$        &  $9$      & $12$       & $15$         \\\hline
QADMM-color    & Time(s) &52.0266  &  28.9591   & 32.5345   &  47.9965   & 30.4193      \\
               & Rate    &22.85$\%$&   49.59$\%$&  63.69$\%$&   68.74$\%$&  72.09$\%$   \\\hline

RADMM-color    & Time(s) &6.2049   &   3.9847   &  6.6568   &   7.5933   &  4.2583      \\
               & Rate    &14.36$\%$&   49.59$\%$&  60.34$\%$&   67.21$\%$&  71.73$\%$   \\\hline

RADMM-gray     & Time(s) &3.0223   &   2.0628   &  3.3289   &   2.9618   &  2.0643      \\
               & Rate    &4.70$\%$ &   37.85$\%$&  56.10$\%$&   66.31    &  68.93$\%$   \\\hline

QPCA-color     & Time(s) &96.3664  &  69.5055   & 84.6165   & 134.0832   &100.5329      \\
               & Rate    &36.04$\%$&   55.47$\%$&  61.52$\%$&   63.60$\%$&  66.21$\%$   \\\hline
\end{tabular}
\end{center}\label{tab11}
\end{table}

%\vspace{0.6cm}

\begin{table}[!ht]
\caption{Numerical results of Example \ref{ex6.2} $(\eta=32)$.}
\begin{center} \footnotesize
\begin{tabular}{|c|c|c|c|c|c|c| } \hline
Method         &$l$      &  $3$    & $6$       &  $9$      & $12$      & $15$       \\\hline
QADMM-color    & Time(s) &22.4030  &  22.6615  &  22.8519  &  22.9498  &  23.6767   \\
               & Rate    &24.88$\%$&  50.57$\%$&  69.76$\%$&  72.68$\%$&  76.42$\%$ \\\hline

RADMM-color    & Time(s) &2.7697   &  2.7544   &  2.7980   &  2.7596   &  2.9234    \\
               & Rate    &15.45$\%$&  44.55$\%$&  67.80$\%$&  72.36$\%$&  74.63$\%$ \\\hline

RADMM-gray     & Time(s) &1.3816   &  1.3950   &  1.4634   &  1.4897   &  1.4620    \\
               & Rate    &5.69$\%$ &  32.68$\%$&  56.91$\%$&  68.29$\%$&  71.87$\%$ \\\hline

QPCA-color     & Time(s) &67.2749  & 72.5781   & 95.2822   &147.7504   &104.1266    \\
               & Rate    &39.67$\%$&  60.49$\%$&  66.83$\%$&  69.11$\%$&  70.41$\%$ \\\hline
\end{tabular}
\end{center}\label{tab12}
\end{table}

For Example \ref{ex6.2}, QADMM-color, RADMM-color, RADMM-gray algorithms  run 4 iterations and
the accuracy rates as well as the CPU time are presented on   Tables \ref{tab9}-\ref{tab12} for different dimensions $l$ and  the random numbers
$\eta$.

\medskip

\begin{figure}[!ht]
\begin{center}
\includegraphics[height=0.35\linewidth,width=1.0\linewidth]{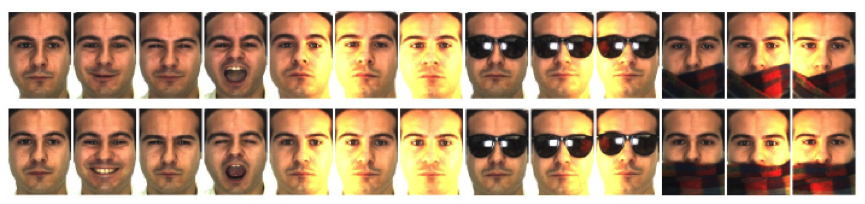}
\end{center}
\caption{Original  color face images   from AR face database. }\label{Fig7}
\end{figure}

\begin{example}\label{ex6.1}
In this experiment, we focus on the famous AR face database \cite{data_AR}.
We  select randomly $100$  individuals  $($$50$ males and $50$ females$)$.
Each individual contains $26$ different color face images  with $165\times 120$ pixels, including frontal views of with different facial
expressions, lighting conditions and occlusions.
The training set is randomly selected with the  random number $\lambda(=14,16,18,20)$
and the remaining as the testing set.
All the color face images are resized to $80\times 60$ pixels.
Fig.~\ref{Fig7} presents the faces from one of the male individuals.
\end{example}

\medskip

For Example \ref{ex6.1}, the QADMM-color, RADMM-color and  RADMM-gray algorithms run 4 iterations and
the accuracy rates as well as the CPU time are presented on   Tables \ref{tab5}-\ref{tab8} for different dimensions $l$ and the values $\lambda$.

\begin{table}[!ht]
\caption{Numerical results of Example \ref{ex6.1} $(\lambda=14)$.}
\begin{center} \footnotesize
\begin{tabular}{|c|c|c|c|c|c|c| } \hline
Method         &$l$      &  $5$    & $10$      &  $15$     & $20$      & $25$       \\\hline
QADMM-color    & Time(s) &8.3088   &  8.4811   & 8.7379    &  8.9954   & 9.0350     \\
               & Rate    &30.58$\%$&  42.67$\%$&  57.00$\%$&  64.58$\%$& 68.08$\%$  \\\hline

RADMM-color    & Time(s) &2.3156   &  2.3267   &  2.3678   &  2.4297   & 2.4932     \\
               & Rate    &27.58$\%$&  41.83$\%$&  53.83$\%$&  61.25$\%$& 66.92$\%$  \\\hline

RADMM-gray     & Time(s) &1.0397   &  1.0343   &  1.0205   &  1.0834   & 1.0727     \\
               & Rate    &18.92$\%$&  33.75$\%$&  45.92$\%$&  54.92$\%$& 59.67$\%$  \\\hline

QPCA-color     & Time(s) &13.4900  & 18.8405   & 21.8072   & 27.3111   & 32.1785    \\
               & Rate    &33.33$\%$&  43.42$\%$&  46.75$\%$&  50.58$\%$& 53.50$\%$  \\\hline

\end{tabular}
\end{center}\label{tab5}
\end{table}

\begin{table}[!ht]
\caption{Numerical results of Example \ref{ex6.1} $(\lambda=16)$.}
\begin{center} \footnotesize
\begin{tabular}{|c|c|c|c|c|c|c| } \hline
Method         &$l$      &  $5$    & $10$      &  $15$     & $20$      & $25$        \\\hline
QADMM-color    & Time(s) &8.9551   &  8.9055   &  9.1517   &  9.3072   &  9.6715     \\
               & Rate    &39.20$\%$&  58.50$\%$&  65.60$\%$&  70.40$\%$&  74.40$\%$  \\\hline

RADMM-color    & Time(s) &1.9035   &  1.9908   &  1.9718   &  2.0098   &  2.1245      \\
               & Rate    &34.30$\%$&  57.10$\%$&  60.60$\%$&  68.20$\%$&  72.70$\%$   \\\hline

RADMM-gray     & Time(s) &0.9110   &  0.9376   &  0.9309   &  0.9337   &  0.9478      \\
               & Rate    &24.70$\%$&  47.40$\%$&  56.00$\%$&  62.50$\%$&  68.60$\%$   \\\hline

QPCA-color     & Time(s) &13.1535  & 19.2302   & 21.6543   & 26.7626   & 31.7443     \\
               & Rate    &45.70$\%$&  58.80$\%$&  63.90$\%$&  67.40$\%$&  68.40$\%$   \\\hline
\end{tabular}
\end{center}\label{tab6}
\end{table}

\begin{table}[!ht]
\caption{Numerical results of Example \ref{ex6.1} $(\lambda=18)$.}
\begin{center} \footnotesize
\begin{tabular}{|c|c|c|c|c|c|c| } \hline
Method         &$l$      &  $5$    & $10$      &  $15$     & $20$      & $25$     \\\hline
QADMM-color    & Time(s) &8.0949   &  8.5623   &  8.9744   &  8.7200   &  9.6274 \\
               & Rate    &25.87$\%$&  43.13$\%$&  50.50$\%$&  58.25$\%$&  65.50$\%$  \\\hline

RADMM-color    & Time(s) &1.8528   &  1.8479   &  1.9877   &  1.9465   &  2.1142 \\
               & Rate    &21.75$\%$&  42.63$\%$&  49.75$\%$&  56.25$\%$&  62.50$\%$ \\\hline

RADMM-gray     & Time(s) &0.8996   &  0.9225   &  1.2338   &  0.8630   &  0.9577  \\
               & Rate    &14.62$\%$&  34.25$\%$&  44.12$\%$&  51.00$\%$&  59.25$\%$  \\\hline

QPCA-color     & Time(s) &13.4892  & 18.9881   & 22.0845   & 27.9783   & 32.1396 \\
               & Rate    &33.62$\%$&  46.37$\%$&  49.13$\%$&  51.50$\%$&  52.88$\%$ \\\hline
\end{tabular}
\end{center}\label{tab7}
\end{table}

\begin{table}[!ht]
\caption{Numerical results of Example \ref{ex6.1} $(\lambda=20)$.}
\begin{center} \footnotesize
\begin{tabular}{|c|c|c|c|c|c|c| } \hline
Method         &$l$      &  $5$    & $10$      &  $15$     & $20$      & $25$     \\\hline
QADMM-color    & Time(s) &9.0368   &  9.1079   &  9.8130   &  9.9078   &   10.2441 \\
               & Rate    &31.00$\%$&  51.50$\%$&  65.67$\%$&  81.50$\%$&   84.17$\%$  \\\hline

RADMM-color    & Time(s) &1.5250   &  1.5520   &  1.5598   &  1.6164   &   1.7145 \\
               & Rate    &30.83$\%$&  50.17$\%$&  62.83$\%$&  78.00$\%$&   82.50$\%$ \\\hline

RADMM-gray     & Time(s) &0.7205   &  0.7291   &  0.7306   &  0.8068   &   0.8167 \\
               & Rate    &17.67$\%$&  40.50$\%$&  52.33$\%$&  59.83$\%$&   69.17$\%$  \\\hline

QPCA-color     & Time(s) &13.9676  & 18.7256   & 21.7172   & 27.5337   &  33.5137  \\
               & Rate    &34.83$\%$&  44.67$\%$&  50.83$\%$&  54.67$\%$&   55.67$\%$
  \\\hline
\end{tabular}
\end{center}\label{tab8}
\end{table}

From the results of Examples \ref{ex6.2} and \ref{ex6.1}, it indicates the following facts.
\begin{itemize}
\item
The results of Examples  \ref{ex6.2} and \ref{ex6.1}  exhibit the recognition performance and  robustness of  QADMM-color algorithm,
which has the better accuracy rate and less CPU time than QPCA-color algorithm  for the  larger value $l$.
This indicates that QADMM-color algorithm has a good numerical performance.

\item
QPCA-color algorithm costs the most CPU time even though we use the fast  Lanczos method.
This is due to that we need to compute the first $l$ largest eigenvalues and corresponding eigenvectors of the covariance matrix from the training set,
which is a $2500\times 2500$ quaternion matrix for Example \ref{ex6.2} or a $4800\times 4800$ quaternion matrix for Example \ref{ex6.1} .

\item
When the value $l$  is small,  the accuracy rates of QADMM-color and QPCA-color are much higher than RADMM-color and  RADMM-gray.
This indicates that the algorithms based on the quaternion are better than encoded three color channels of color images and single channel of gray level images.
In other word, color images represented the pure quaternion matrices can explore the information between color channels.
In addition, with the increase of the dimension $l$, the accuracy rate of face recognition also increases.

\end{itemize}

\section{Concluding remarks}\label{sec6}

We introduced the QNQMF model and proposed
two solving numerical algorithms.
Moreover, we explored the advantages of QNQMF for color image  reconstruction and color  face recognition.
From these numerical results, it can see that
QNQMF has  simple form, good interpretability and small storage space.
And the algorithms encoded on the quaternion perform better than the algorithms
encoded on the red, green and blue channels.
In particular, the QNQMF model has a better face recognition ability  and robustness than
real NMF model encoded  separately on color channels of color images and on single channel of gray level images for the same data.

\clearpage
\textbf{Acknowledgments}
The authors are grateful for the Editor-in-Chief, Associate Editor, and Reviewers for their valuable comments and insightful suggestions that helped to improve this research significantly.

\textbf{Author Contributions} The authors have contributed equally in this scholarly work.

\textbf{Funding}
This research is supported by the National Natural Science Foundation of China  grants 62105064, 41725017, 12171210, 12090011 and 11901098;
the Major Projects of Universities in Jiangsu Province of China grants  21KJA110001;
and the Natural Science Foundation of Fujian Province of China grants 2020J05034.

\bigskip

\appendix

\section{ The real representation method}  \label{sec:appendixB}
In this appendix, we analysis the real representation method.

Remark that (\ref{eqb1}) is equivalent to
\begin{equation}\label{eqb2}
\Upsilon_\mathbf{X}=\Upsilon_\mathbf{W}\Upsilon_\mathbf{H},
\end{equation}
or
\begin{equation}\label{eqb3}
\left\{
  \begin{array}{c}
X_0=W_0H_0-W_1H_1-W_2H_2-W_3H_3, \\
X_1=W_0H_1+W_1H_0+W_2H_3-W_3H_2, \\
X_2=W_0H_2-W_1H_3+W_2H_0+W_3H_1, \\
X_3=W_0H_3+W_1H_2-W_2H_1+W_3H_0 \\
  \end{array}
\right.
\end{equation}
with $W_1,W_2,W_3\geq 0$ and $H_1,H_2,H_3\geq 0$.

For the equivalent problem (\ref{eqb2}), it is quite expansive to solve
as the dimension-expanding problem caused by the real counterpart method.
Meanwhile, it is  different from  NMF (\ref{eqa1});
thus those numerical algorithms proposed for NMF will not directly applied to solve (\ref{eqb2}).
And for the  problem (\ref{eqb3}), it will be difficult to comprehend adequately the structure of quaternion quasi non-negative matrices $\mathbf{W}$ and $\mathbf{H}$ if $W_0,W_1,W_2,W_3$ and $H_0,H_1,H_2,H_3$
are as  separate variables.

In fact,  the optimization problem (\ref{eqb4}) is equivalent to
\begin{equation}\label{eqJia1}
\begin{array}{ll}
\min &f(\mathbf{W},\mathbf{H})=\|X_0-(W_0H_0-W_1H_1-W_2H_2-W_3H_3)\|_F^2\\
&\qquad\qquad~~ +\|X_1-(W_0H_1+W_1H_0+W_2H_3-W_3H_2)\|_F^2\\
&\qquad\qquad~~ +\|X_2-(W_0H_2-W_1H_3+W_2H_0+W_3H_1)\|_F^2\\
&\qquad\qquad~~ +\|X_3-(W_0H_3+W_1H_2-W_2H_1+W_3H_0)\|_F^2, \\
\mathrm{s.t.} &W_1\geq 0,~W_2\geq 0,~W_3\geq 0,~H_1\geq 0,~H_2\geq 0,~H_3\geq 0.
\end{array}
\end{equation}

Now, we consider  the Karush-Kuhn-Tucker (KKT) optimality conditions of the optimization problem (\ref{eqJia1}).
Let $y=(y_1^T,y_2^T,\ldots,y_8^T)^T\in \mathbb{R}^{4l(m+n)}$ with
$$
\begin{array}{llll}
y_1=\mathrm{vec}(W_0), & y_2=\mathrm{vec}(W_1), & y_3=\mathrm{vec}(W_2), & y_4=\mathrm{vec}(W_3),\\
y_5=\mathrm{vec}(H_0), & y_6=\mathrm{vec}(H_1), & y_7=\mathrm{vec}(H_2), & y_8=\mathrm{vec}(H_3),\\
\end{array}
$$
and
$$
\begin{array}{lll}
G_{1,s,t}(y)=(W_1)_{st}, & G_{2,s,t}(y)=(W_2)_{st}, & G_{3,s,t}(y)=(W_3)_{st},\\
J_{1,t,k}(y)=(H_1)_{tk}, & J_{2,t,k}(y)=(H_2)_{tk}, & J_{3,t,k}(y)=(H_3)_{tk}\\
\end{array}
$$
for $s=1,\ldots,m$, $t=1,\ldots,l$ and $k=1,\ldots,n$.

Suppose that $\widehat{y}$ is a local minimum of the problem (\ref{eqJia1}). Denote
$$\mathbb{I}_G(\widehat{y})=\{(a,s,t)\,|\,G_{a,s,t}(\widehat{y})=0\},~~\mathbb{I}_J(\widehat{y})=\{(b,t,k)\,|\,J_{b,t,k}(\widehat{y})=0\}.$$
It is obvious that $f$, $G_{a,s,t}(a=1,2,3;s=1,\ldots,m;t=1,\ldots,l)$ and $J_{b,t,k}(b=1,2,3;t=1,\ldots,l;k=1,\ldots,n)$ are continuously differentiable functions over   $\mathbb{R}^{4l(m+n)}$.
And it is easy to verify that the gradients of the active constraints
$$\{\nabla G_{a,s,t}(\widehat{y})\}_{(a,s,t)\in \mathbb{I}_G(\widehat{y})}\cup \{\nabla J_{b,t,k}(\widehat{y})\}_{(b,t,k)\in \mathbb{I}_J(\widehat{y})}$$
are linearly independent.
Then, there exist multipliers  $c_{a,s,t}(a=1,2,3;s=1,\ldots,m;t=1,\ldots,l)$ and $d_{b,t,k}(b=1,2,3;t=1,\ldots,l;k=1,\ldots,n)$ such that
\begin{equation}\label{eqJia2}
\left\{
  \begin{array}{l}
\nabla f(\widehat{y})-\sum_{a,s,t}c_{a,s,t} \nabla G_{a,s,t}(\widehat{y})-\sum_{b,t,k}d_{b,t,k} \nabla J_{b,t,k}(\widehat{y})=0,\\
G_{a,s,t}(\widehat{y})\geq 0,~~~~c_{a,s,t}\geq 0,~~~~c_{a,s,t}G_{a,s,t}(\widehat{y})=0,  \\
J_{b,t,k}(\widehat{y})\geq 0,~~~~~d_{b,t,k}\geq 0,~~~~d_{b,t,k}J_{b,t,k}(\widehat{y})=0,
  \end{array}
\right.
\end{equation}
where $a,b=1,2,3;$ $s=1,\ldots,m;$ $t=1,\ldots,l$ and $k=1,\ldots,n$.
Let the matrices $C_a=(c_{a,s,t})\in \mathbb{R}^{m\times l}$ with $a=1,2,3$ and  $D_b=(d_{b,t,k})\in \mathbb{R}^{l\times n}$ with $b=1,2,3$.
Then, if $(\widehat{\mathbf{W}},\widehat{\mathbf{H}})$  corresponding with $\widehat{y}$ is a local minimum of the problem (\ref{eqJia1}), from (\ref{eqJia2}),
we have
\begin{equation}\label{eqJia3}
\left\{
  \begin{array}{l}
\nabla_{W_0} f(\widehat{\mathbf{W}},\widehat{\mathbf{H}})=0,~~\nabla_{H_0} f(\widehat{\mathbf{W}},\widehat{\mathbf{H}})=0,\\
\nabla_{W_1} f(\widehat{\mathbf{W}},\widehat{\mathbf{H}})-C_1=0,~~W_1\geq 0,~~C_1\geq 0,~~ W_1\odot C_1=0,\\
\nabla_{W_2} f(\widehat{\mathbf{W}},\widehat{\mathbf{H}})-C_2=0,~~W_2\geq 0,~~C_2\geq 0,~~ W_2\odot C_2=0,\\
\nabla_{W_3} f(\widehat{\mathbf{W}},\widehat{\mathbf{H}})-C_3=0,~~W_3\geq 0,~~C_3\geq 0,~~ W_3\odot C_3=0,\\
\nabla_{H_1} f(\widehat{\mathbf{W}},\widehat{\mathbf{H}})-D_1=0,~~H_1\geq 0,~~D_1\geq 0,~~ H_1\odot D_1=0,\\
\nabla_{H_2} f(\widehat{\mathbf{W}},\widehat{\mathbf{H}})-D_2=0,~~H_2\geq 0,~~D_2\geq 0,~~ H_2\odot D_2=0,\\
\nabla_{H_3} f(\widehat{\mathbf{W}},\widehat{\mathbf{H}})-D_3=0,~~H_3\geq 0,~~D_3\geq 0,~~ H_3\odot D_3=0.
  \end{array}
\right.
\end{equation}
According to (\ref{eqJia3}), we can obtain the KKT  optimality conditions of the problem (\ref{eqb4}), that is (\ref{eqb6}).

\section{Proofs of Lemmas \ref{lem3-2}, \ref{lem3-3} and \ref{lem3-5}} \label{sec:appendixA}
In this appendix, we present the proofs of lemmas in Section \ref{sec3}.

Define the matrix set
$$\mathbb{R}_\dag^{m\times 4n}:=\{R=(R_0,R_1,R_2,R_3)\in\mathbb{R}^{m\times 4n}\,|\,
R_0\in \mathbb{R}^{m\times n}, R_1,R_2,R_3\in \mathbb{R}_+^{m\times n}\},$$
where $\mathbb{R}_+^{m\times n}$ is the non-negative set on  the real field.
Hence, the sets $\mathbb{Q}_\dag^{m\times n}$ and $\mathbb{R}_\dag^{m\times 4n}$ are isomorphic.

It is obvious that $\mathbb{R}_\dag^{m\times 4n}$ is a nonempty closed convex set,
and let $\mathcal{P}_{\mathbb{R}_\dag^{m\times 4n}}$ be the projection into $\mathbb{R}_\dag^{m\times 4n}$, which is defined by
$$\mathcal{P}_{\mathbb{R}_\dag^{m\times 4n}}(R):=\mathrm{argmin}\, \|Y-R\|_F,\quad \forall\,
Y\in \mathbb{R}_\dag^{m\times 4n},$$
then we can obtain that
$$\mathcal{P}_{\mathbb{R}_\dag^{m\times n}}(R)=
(R_0,~\mathcal{P}_+(R_1),~\mathcal{P}_+(R_2),~\mathcal{P}_+(R_3)),
$$
where $\mathcal{P}_+(R_s)=\max(R_s,0)$, $s=1,2,3$.

\bigskip

{\bf Proof of Lemma \ref{lem3-2}.} Let $\mathbf{X}=X_0 + X_1{\bf i} + X_2{\bf j} + X_3{\bf k}$,
$\mathbf{W}=W_0 + W_1{\bf i} + W_2{\bf j} + W_3{\bf k}$
and
$\mathbf{H}=H_0 + H_1{\bf i} + H_2{\bf j} + H_3{\bf k}$.
Since
\begin{eqnarray}
\mathbf{W}\mathbf{H}=&&(W_0H_0-W_1H_1-W_2H_2-W_3H_3)+
(W_0H_1+W_1H_0+W_2H_3-W_3H_2){\bf i}\nonumber\\
&&+(W_0H_2-W_1H_3+W_2H_0+W_3H_1){\bf j}+
(W_0H_3+W_1H_2-W_2H_1+W_3H_0){\bf k}\nonumber\\
:=&&M_0+ M_1{\bf i} + M_2{\bf j} + M_3{\bf k}\nonumber
\end{eqnarray}
and
$$\frac{1}{2}\|\mathbf{X}-\mathbf{W}\mathbf{H}\|_F^2=\frac{1}{2}(\|X_0-M_0\|_F^2+\|X_1-M_1\|_F^2+\|X_2-M_2\|_F^2
+\|X_3-M_3\|_F^2),$$
according to (\ref{eqb5W}), we can get that
\begin{align*}
&\nabla_\mathbf{W} f(\mathbf{W},\mathbf{H})=\frac{\partial f(\mathbf{W},\mathbf{H})}{\partial W_0}+\frac{\partial f(\mathbf{W},\mathbf{H})}{\partial W_1}{\bf i}
+\frac{\partial f(\mathbf{W},\mathbf{H})}{\partial W_2}{\bf j}+\frac{\partial f(\mathbf{W},\mathbf{H})}{\partial W_3}{\bf k}\nonumber\\
=&(X_0-M_0)(-H_0^T)+(X_1-M_1)(-H_1^T)+(X_2-M_2)(-H_2^T)+(X_3-M_3)(-H_3^T)\nonumber\\
&+[(X_0-M_0)(H_1^T)+(X_1-M_1)(-H_0^T)+(X_2-M_2)(H_3^T)+(X_3-M_3)(-H_2^T)]{\bf i}\nonumber\\
&+[(X_0-M_0)(H_2^T)+(X_1-M_1)(-H_3^T)+(X_2-M_2)(-H_0^T)+(X_3-M_3)(H_1^T)]{\bf j}\nonumber\\
&+[(X_0-M_0)(H_3^T)+(X_1-M_1)(H_2^T)+(X_2-M_2)(-H_1^T)+(X_3-M_3)(-H_0^T)]{\bf k}\nonumber\\
=&-(\mathbf{X}-\mathbf{W}\mathbf{H})\mathbf{H}^*.\nonumber
\end{align*}

Similarly, we can get that $\nabla_{\mathbf{H}} f(\mathbf{W},\mathbf{H})=-\mathbf{W}^*(\mathbf{X}-\mathbf{W}\mathbf{H})$.
\hfill
\endproof

\bigskip

{\bf Proof of Lemma \ref{lem3-3}.} It is obvious that if the point $(\widehat{\mathbf{W}},\widehat{\mathbf{H}})$ satisfies (\ref{eqb6}), then it meets (\ref{eqb7}).
We will show next that conversely: if (\ref{eqb7}) holds, then the point $(\widehat{\mathbf{W}},\widehat{\mathbf{H}})$  also meets  (\ref{eqb6}).

Case 1: if $\mathrm{Re}\,\nabla_\mathbf{W} f(\widehat{\mathbf{W}},\widehat{\mathbf{H}})\neq0$, then there exist $s_0$ and $t_0$ such that
$a_0:=(\mathrm{Re}\,\nabla_\mathbf{W} f(\widehat{\mathbf{W}},\widehat{\mathbf{H}}))_{s_0,t_0}\neq 0$. Let
$$\mathbf{Y}_{st}=\left\{
           \begin{array}{ll}
             \widehat{\mathbf{W}}_{st}-a_0, & {if}~s=s_0,~t=t_0, \\
             \widehat{\mathbf{W}}_{st},     & {others}. \\
           \end{array}
         \right.
$$
As $\widehat{\mathbf{W}}\in   \mathbb{Q}_\dag^{m\times l}$, thus we have $\mathbf{Y}\in\mathbb{Q}_\dag^{m\times l}$ and
$$\mathrm{Re}\, [\langle \nabla_\mathbf{W} f(\widehat{\mathbf{W}},\widehat{\mathbf{H}}),~\mathbf{Y}-\widehat{\mathbf{W}}\rangle]=-a_0^2<0,$$
which is contradict with the first inequality of (\ref{eqb7}).

Case 2: if $\nabla_\mathbf{W} f(\widehat{\mathbf{W}},\widehat{\mathbf{H}})\not\in \mathbb{Q}_\dag^{m\times l}$, there exist
$s_0$ and $t_0$ such that one of the following three inequalities
$$\left\{
    \begin{array}{l}
(\mathrm{Im}_i\,\nabla_\mathbf{W} f(\widehat{\mathbf{W}},\widehat{\mathbf{H}}))_{s_0,t_0}<0,\\
(\mathrm{Im}_j\,\nabla_\mathbf{W} f(\widehat{\mathbf{W}},\widehat{\mathbf{H}}))_{s_0,t_0}<0,\\
(\mathrm{Im}_k\,\nabla_\mathbf{W} f(\widehat{\mathbf{W}},\widehat{\mathbf{H}}))_{s_0,t_0}<0\\
    \end{array}
  \right.
$$
holds. Without loss of generality, assume that $a_1:=(\mathrm{Im}_i\,\nabla_\mathbf{W } f(\widehat{\mathbf{W}},\widehat{\mathbf{H}}))_{s_0,t_0}<0$.
Let
$$\mathbf{Y}_{st}=\left\{
           \begin{array}{ll}
             \widehat{\mathbf{W}}_{st}-a_1\mathbf{i}, &{if}~ s=s_0,~t=t_0, \\
             \widehat{\mathbf{W}}_{st},& {others}. \\
           \end{array}
         \right.
$$
Then, $(\mathrm{Im}_i\,\mathbf{Y})_{s_0,t_0}=(\mathrm{Im}_i\,\widehat{\mathbf{W}})_{s_0,t_0}-a_1>0$. Thus, we have
$\mathbf{Y}\in\mathbb{Q}_\dag^{m\times l}$ and
$$\mathrm{Re}\, [\langle \nabla_\mathbf{W} f(\widehat{\mathbf{W}},\widehat{\mathbf{H}}),~\mathbf{Y}-\widehat{\mathbf{W}}\rangle]=-a_1^2<0,$$
which is contradict with the first inequality of (\ref{eqb7}).

Case 3: if $\mathrm{Im}_i\,\widehat{\mathbf{W}}\odot \mathrm{Im}_i\,\nabla_\mathbf{W} f(\widehat{\mathbf{W}},\widehat{\mathbf{H}})\neq0,$ there exist
$s_0$ and $t_0$ such that
$$\widehat{w}_1:=(\mathrm{Im}_i\,\widehat{\mathbf{W}})_{s_0,t_0}>0,\quad
a_1:=(\mathrm{Im}_i\,\nabla_\mathbf{W} f(\widehat{\mathbf{W}},\widehat{\mathbf{H}}))_{s_0,t_0}>0.$$
Let
$$\mathbf{Y}_{st}=\left\{
           \begin{array}{ll}
             0,               &{if}~ s=s_0,~t=t_0, \\
             \widehat{\mathbf{W}}_{st},& {others}. \\
           \end{array}
         \right.
$$
Then, we have
$\mathbf{Y}\in\mathbb{Q}_\dag^{m\times l}$ and
$$\mathrm{Re}\, [\langle \nabla_\mathbf{W} f(\widehat{\mathbf{W}},\widehat{\mathbf{H}}),~\mathbf{Y}-\widehat{\mathbf{W}}\rangle]=-\widehat{w}_1a_1<0,$$
which is contradict with the first inequality of (\ref{eqb7}).

If $\mathrm{Im}_j\,\widehat{\mathbf{W}}\odot \mathrm{Im}_j\,\nabla_\mathbf{W} f(\widehat{\mathbf{W}},\widehat{\mathbf{H}})\neq0$
and $\mathrm{Im}_k\,\widehat{\mathbf{W}}\odot \mathrm{Im}_k\,\nabla_\mathbf{W} f(\widehat{\mathbf{W}},\widehat{\mathbf{H}})\neq0$,
we can also show that they are contradict with the first inequality of (\ref{eqb7}).
Hence, if the first inequality of (\ref{eqb7}) holds if and only if $\widehat{\mathbf{W}}$ satisfies (\ref{eqb6}).

Similarly, it can verify that the second inequality of (\ref{eqb7}) holds if and only if $\widehat{\mathbf{H}}$ satisfies (\ref{eqb6}).
\hfill\endproof

\bigskip

{\bf Proof of Lemma \ref{lem3-5}.}
Let $\mathbf{Y}=Y_0 + Y_1{\bf i} + Y_2{\bf j} + Y_3{\bf k}\in \mathbb{Q}^{m\times n}$ and
$\mathbf{Z}=Z_0 + Z_1{\bf i} + Z_2{\bf j} + Z_3{\bf k}\in \mathbb{Q}^{m\times n}$.

We prove the result (1).
If $\mathbf{Z}\in \mathbb{Q}_\dag^{m\times n}$, it follows that
$$\mathrm{Re}\,[\langle \mathcal{P}_{\mathbb{Q}_\dag^{m\times n}}(\mathbf{Y})-\mathbf{Y},~
\mathbf{Z}-\mathcal{P}_{\mathbb{Q}_\dag^{m\times n}}(\mathbf{Y})\rangle ]=
\sum_{s=1}^3 \langle \mathcal{P}_+(Y_s) -Y_s ,~ Z_s-\mathcal{P}_+(Y_s)\rangle \geq 0,
$$
which is based on the fact $\langle \mathcal{P}_+(a) -a ,~ b-\mathcal{P}_+(a)\rangle\geq 0$ for any $a\in \mathbb{R}$ and $b\in \mathbb{R}_+$.

We prove the result (2).
It follows that
$$
\begin{array}{ll}
&\mathrm{Re}\,[\langle \mathcal{P}_{\mathbb{Q}_\dag^{m\times n}}(\mathbf{Y})-
\mathcal{P}_{\mathbb{Q}_\dag^{m\times n}}(\mathbf{Z}),\mathbf{Y}-\mathbf{Z}\rangle ]\\
&=\langle Y_0-Z_0, Y_0-Z_0 \rangle+
\sum_{s=1}^3 \langle \mathcal{P}_+(Y_s) -\mathcal{P}_+(Z_s) ,~ Y_s-Z_s\rangle\geq 0,
\end{array}
$$
which is based on the fact $\langle \mathcal{P}_+(a) -\mathcal{P}_+(b),~ a-b\rangle\geq 0$ for any $a,b\in \mathbb{R}$.
It is obvious that $\mathcal{P}_{\mathbb{Q}_\dag^{m\times n}}(\mathbf{Y})\neq \mathcal{P}_{\mathbb{Q}_\dag^{m\times n}}(\mathbf{Z})$, then the strict inequality holds.

We prove the result (3).
It follows that
$$
\begin{array}{ll}
&\|\mathcal{P}_{\mathbb{Q}_\dag^{m\times n}}(\mathbf{Y})-\mathcal{P}_{\mathbb{Q}_\dag^{m\times n}}(\mathbf{Z})\|_F^2=\|Y_0-Z_0\|_F^2+\sum_{s=1}^3\|\mathcal{P}_+(Y_s)-\mathcal{P}_+(Z_s)\|_F^2\\
&\leq \|Y_0-Z_0\|_F^2+\sum_{s=1}^3\|Y_s-Z_s\|_F^2=\|\mathbf{Y}-\mathbf{Z}\|_F^2,
\end{array}
$$
which is based on the fact $|\mathcal{P}_+(a)-\mathcal{P}_+(b)|\leq |a-b|$ for any  $a,b\in \mathbb{R}$.
\hfill\endproof

\bigskip


\begin{thebibliography}{10}

\bibitem{NMF_ALS2}
Ang, A.M.S., Gillis, N.:
Accelerating nonnegative matrix factorization algorithms using extrapolation.
Neural Comput.
\textbf{31}(2), 417-439 (2019)


\bibitem{D1976}
Bertsekas, D.P.:
On the Goldstein-Levitin-Polyak gradient projection method.
IEEE Trans. Automat. Contr.
\textbf{21}(2), 174-184 (1976)


\bibitem{ADMMBoyd2011}
Boyd, S., Parikh, N., Chu, E., Peleato, B., Eckstein, J.:
Distributed optimization and statistical learning via the alternating direction
method of multipliers.
Found. Trends Mach. Learn.
\textbf{3}(1), 1-122 (2011)

\bibitem{Calamai1987}
Calamai, P.P., Mor$\acute{e}$, J.J.:
Projected gradient methods for linearly constrained problems.
Math. Program.
\textbf{39}, 93-116 (1987)



\bibitem{Q_watermarking}
Chen, Y., Jia, Z.G., Peng, Y., Zhang, D.:
A new structure-preserving quaternion QR decomposition method for color image blind watermarking.
Signal Process.
\textbf{185}, 108088 (2021)

\bibitem{Chen2020}
Chen, Y.N., Qi, L.Q., Zhang, X.Z., Xu, Y.W.:
A low rank quaternion decomposition algorithm and its application
in color image inpainting.
arXiv:2009.12203 (2020)



\bibitem{NMF_RALS}
Cichocki, A., Zdunek, R.:
Regularized alternating least squares algorithms for non-negative matrix/tensor factorization.
Advances in Int. Symposium on Neural Networks, Springer, Berlin, Heidelberg
793-802 (2007)

\bibitem{Flamant2020}
Flamant, J., Miron, S., Brie, D.:
Quaternion non-negative matrix factorization: Definition, uniqueness, and algorithm.
IEEE Trans. Signal Process.
\textbf{68}, 1870-1883 (2020)



\bibitem{Gafni1982}
Gafni, E.M., Bertsekas, D.P.:
Convergence of a gradient projection method.
Report LIDS-P-1201, Lab. for Info. and Dec. Systems, M.I.T. (1982)


\bibitem{ADMMGlowinski1975}
Glowinski, R., Marrocco, A.:
Sur l'approximation
par elements finis d'ordre un, et la resolution par
penalisation-dualite d’une classe de problemes de
Dirichlet nonlineaires.
ESAIM: Mathematical Moddelling and Numerical Analysis  Modlisation Mathmatique et Analyse Numrique
\textbf{9},  41-76  (1975)



\bibitem{NMF_Newton1}
Gong, P.H., Zhang, C.S.:
Efficient nonnegative matrix factorization via projected Newton method.
Pattern Recogn.
\textbf{45}(9), 3557-3565 (2012)




\bibitem{NMF_NeNMF}
Guan, N.Y., Tao, D.C., Luo, Z.G., Yuan, B.:
NeNMF: an optimal gradient method for nonnegative matrix factorization.
IEEE Trans. Signal Process.
\textbf{60}, 2882-2898 (2012)



\bibitem{NMF_ADMM2}
Hajinezhad, D., Chang, T.H., Wang, X.F.,  Shi, Q.J., Hong, M.Y.:
Nonnegative matrix factorization using ADMM: Algorithm and convergence analysis.
In International Conference on Acoustics, Speech and Signal Processing (ICASSP),
IEEE, 4742-4746 (2016)


\bibitem{Q_Hamilton}
Hamilton, W.R.:
Elements of Quaternions.
Longmans Green, London  (1866)


\bibitem{bu3}
Huang, C.Y., Fang, Y.Y., Wu, T.T., Zeng, T.Y., Zeng, Y.H.:
Quaternion screened Poisson equation for low-light image enhancement.
IEEE Signal Process. Lett.
\textbf{29}, 1417-1421 (2022)


\bibitem{bu2}
Huang, C.Y., Li, Z., Liu, Y.B., Wu, T.T., Zeng, T.Y.:
Quaternion-based weighted nuclear norm minimization for color image restoration.
Pattern Recogn.
\textbf{128}, 108665 (2022)


\bibitem{Jia2019}
Jia, Z.G.:
The Eigenvalue Problem of Quaternion Matrix: Structure-Preserving Algorithms
and Applications.
Science Press, Beijing (2019)

\bibitem{ke3}
Jia, Z.G., Ng, M.K.:
Structure preserving quaternion generalized minimal residual method.
SIAM J. Matrix Anal. Appl.
\textbf{42}(2), 616-634 (2021)


\bibitem{ke0}
Jia, Z.G.,  Jin, Q.Y., Ng, M.K., Zhao, X.L.:
Non-local robust quaternion matrix completion for large-scale color image and video inpainting.
IEEE Trans. Image Process.
\textbf{31}, 3868-3883 (2022)

\bibitem{Q_inpainting1}
Jia, Z.G., Ng, M.K., Song, G.J.:
Robust quaternion matrix completion with applications to image inpainting.
Numer. Linear Algebra Appl.
\textbf{26}(4), e2245 (2019)


\bibitem{QPCA_Lancos}
Jia, Z.G., Ng, M.K., Song, G.J.:
Lanczos method for large-scale quaternion singular value decomposition.
Numer. Algor.
\textbf{82}, 699-717 (2019)


\bibitem{ke5}
Jia, Z.G.,  Ng, M.K., Wang, W.:
Color image restoration by saturation-value total variation.
SIAM J. Imaging Sci.
\textbf{12}(2), 972-1000 (2019)


\bibitem{NMF_ALS1}
Kim, H., Park, H.:
Nonnegative matrix factorization based on alternating nonnegativity constrained least squares and
active set method.
SIAM J. Matrix Anal. Appl.
\textbf{30}(2), 713-730 (2008)


\bibitem{NMF_Lee}
Lee, D.D., Seung, H.S.:
Learning the parts of objects by non-negative matrix factorization.
Nature
\textbf{401}, 788-791 (1999)


\bibitem{Q_watermarking2}
Li, J.Z., Yu, C.Y.,  Gupta, B.B., Ren, X.C.:
Color image watermarking scheme based on quaternion Hadamard transform and Schur decomposition.
Multimed. Tools Appl.
\textbf{77}(4), 4545-4561 (2018)


\bibitem{NMF_Lin}
Lin, C.J.:
Projected gradient methods for nonnegative matrix factorization.
Neural Comput.
\textbf{19}(10), 2756-2779 (2007)


\bibitem{ke2}
Liu, Q.H., Ling, S.T., Jia, Z.G.:
Randomized quaternion singular value decomposition for low-rank matrix approximation.
SIAM J. Sci. Comput.
\textbf{44}(2), A870-A900 (2022)


\bibitem{NMF_constrained}
Lu, X.Q., Wu, H., Yuan, Y.:
Double constrained NMF for hyperspectral unmixing.
IEEE Trans. Geosci. Remote. Sens.
\textbf{52}(5), 2746-2758 (2013)

\bibitem{data_AR}
Martinez, A.M., Benavente, R.:
The AR Face Database. CVC Technical Report 24  (1998)

\bibitem{PT1994}
Paatero, P., Tapper, U.:
Positive matrix factorization: A non-negative
factor model with optimal utilization of error estimates of data values.
Environmetrics
\textbf{5}, 111-126 (1994)


\bibitem{NMF_orthogonal}
Pompili, F., Gillis, N., Absil, P.A., Glineur, F.:
Two algorithms for orthogonal nonnegative matrix factorization with application to clustering.
Neurocomputing
\textbf{141}, 15-25 (2014)

\bibitem{Qi2022}
Qi, L.Q., Luo, Z.Y., Wang Q.W., Zhang, X.Z.:
Quaternion matrix optimization: Motivation and analysis.
J. Optim. Theory Appl.
\textbf{193}(1), 621-648 (2022)


\bibitem{ADMM_color}
Rajapakse, M., Tan, J., Rajapakse, J.C.:
Color channel encoding with NMF for face recognition.
2004 International Conference on Image Processing, IEEE  (2004)


\bibitem{NMF_sparse}
Rapin, J., Bobin, J., Larue, A., Starck, J.L.:
NMF with sparse regularizations in transformed domains.
SIAM J. Imaging Sci.
\textbf{7}(4), 2020-2047 (2014)

\bibitem{ke4}
Song, G.J., Ding, W.Y., Ng, M.K.:
Low rank pure quaternion approximation for pure quaternion matrices.
SIAM J. Matrix Anal. Appl.
\textbf{42}(1), 58-82 (2021)


\bibitem{data_CASIA}
The CASIA 3D Face Database.
http://www.cbsr.ia.ac.cn/english/3dface\%20databases.asp

%\bibitem{data_GT}
%The Georgia Tech face database.
%http://www.anefian.com/research/face\underline{~}reco.htm




\bibitem{bu1}
Wu, T.T., Mao, Z.H., Li, Z.Y., Zeng, Y.H., Zeng, T.Y.:
Efficient color image segmentation via quaternion-based $L_1$/$L_2$ Regularization.
J. Sci. Comput.
\textbf{93}(1), 1-26 (2022)


\bibitem{Q_PCA}
Xiao, X.L.,  Zhou, Y.C.:
Two-dimensional quaternion PCA and sparse PCA.
IEEE Trans. Neur. Net. Lear.
\textbf{30}(7), 2028-2042 (2018)


\bibitem{tensor_NTF}
Xu, Y.Y., Yin, W.T.:
A block coordinate descent method for regularized multiconvex optimization with applications to nonnegative tensor factorization and completion.
SIAM J. Imaging Sci.
\textbf{6}(3), 1758-1789  (2013)


\bibitem{NMF_ADMM1}
Zhang, S.F., Huang, D.Y., Xei, L., Chng, E.S., Li, H.Z., Dong, M.H.:
Non-negative matrix factorization using stable alternating direction method of multipliers for source separation.
In Asia-Pacific Signal and Information Processing Association Annual Summit and Conference (APSIPA), IEEE, 222-228 (2015)


\bibitem{Q_PCA1}
Zhao, M.X., Jia, Z.G., Cai, Y.F., Chen, X., Gong, D.W.:
Advanced variations of two-dimensional principal component analysis for face recognition.
Neurocomputing
\textbf{452}, 653-664 (2021)











\end{thebibliography}
\end{document}